\documentclass[10pt,a4paper]{article}
\usepackage[T1]{fontenc}
\usepackage[utf8]{inputenc}
\usepackage{lmodern}
\usepackage[margin=1in]{geometry}
\usepackage{microtype}
\usepackage{amsmath,amssymb,amsthm}
\usepackage{graphicx}
\usepackage{booktabs,multirow,array,tabularx,enumitem}
\usepackage{xcolor}
\usepackage{algorithm}
\usepackage{algorithmicx}
\usepackage{algpseudocode}
\usepackage{subfig}
\usepackage{float}
\usepackage{natbib}
\usepackage{url}
\usepackage[hidelinks]{hyperref}
\usepackage{cleveref}
\usepackage{tikz}
\usepackage{caption}
\title{Concept-Guided Exploration: Building Persistent, Actionable Scene Graphs}
\author{
Noé Zapata$^{1}$, Gerardo Pérez$^{1}$, Alejandro Torrejón$^{1}$\\
Pedro Núñez$^{1}$, Pablo Bustos$^{1,*}$\\[4pt]
\small $^{1}$RoboLab, Robotics and Artificial Vision, University of Extremadura, 10003 Cáceres, Spain\\
\small \texttt{nzapata@unex.es; gperezgonz@unex.es; atorrejon@unex.es; pnuntru@unex.es; pbustos@unex.es}\\
\small $^{*}$Corresponding author: pbustos@unex.es
}
\date{}

\begin{document}
\maketitle

\begin{abstract}
The perception of 3D space by mobile robots is rapidly moving from flat metric grid representations to hybrid metric-semantic graphs built from human-interpretable concepts. While most approaches first build metric maps and then add semantic layers, we explore an alternative, concept-first architecture in which spatial understanding emerges from asynchronous concept agents that directly instantiate and manage semantic entities. Our robot employs two spatial concepts—room and door—implemented as autonomous processes within a cognitive distributed architecture. These concept agents cooperatively build a shared scene graph representation of indoor layouts through active exploration and incremental validation. The key architectural principle is hierarchical constraint propagation: Room instantiation provides geometric and semantic priors to guide and support door detection within wall boundaries. The resulting structure is maintained by a complementary functional principle based on prediction-matching loops. This approach is designed to yield an actionable, human-interpretable spatial representation without relying on any pre-existing global metric map, supporting scalable operation and persistent, task-relevant understanding in structured indoor environments.
\end{abstract}


%%%%%%%%%%%%%%%%%%%%%%%%%%%%%%%%%%%%%%%%%%

\section{Introduction}

\textcolor{black}{Robots operating in human environments benefit from semantically rich, shared representations of their surroundings}. In recent years, 3D scene graphs (3DSGs) have been used for this purpose, as complex data structures that represent scene objects as nodes and their various relationships as edges \cite{armeni20193dscenegraphstructure}. When used to describe large fragments of space, they are typically organised hierarchically, with higher levels encompassing larger spatial aggregations \cite{hughes2024foundations}. 
The computer vision community has developed many algorithms and DNN models that can be used to detect elements in the scene, their relationships, and even to build the graph directly from complete views of the scene \cite{wu2021scenegraphfusionincremental3dscene}
\cite{huang2018holistic}
\cite{Liu-2018}.
Very complex graphs can also be built in real time using data captured by a human-operated robot, as demonstrated in recent frameworks \cite{rosinol2021kimeraslamspatialperception}\cite{hughes2022hydra}. \textcolor{black}{In most experiments, robots are typically used passively}, in a remote-controlled setup, to capture 3D data in large environments. 

However, if robots are to use 3DSGs in real-time operations, they must build them incrementally, starting from an empty representation when necessary, and proposing actions that improve the current representation by reducing its uncertainty. These actions must coincide with or compete with other requests from other high-priority tasks, forcing the representational system to be opportunistic at its core. Scene graph construction should be driven by the robot's need to understand its world and current task requirements. This opportunistic process may leave objects partially defined for later refinement, requiring the architecture to continuously update its world beliefs. This graph-building loop underlies the control architecture, adapting to ongoing tasks while competing for hardware access.

% In this paper, we explore this view of 3DSG construction, where the robot decides what action to take next to improve or extend the current representation. We assume that the world is composed of rectangular rooms connected by doors. The robot is given a functional definition of the room and door concepts, and the goal is to incrementally build a scene-graph representation of the environment by instantiating these two concepts and relating the resulting instances in geometric and semantic ways. The graph is to be constructed directly from these concepts, avoiding a previous free/occupied grid representation. The robot has a 3D LiDAR that provides a stream of scene-measured points. Finally, we use the CORTEX architecture \cite{bustos_cortex_2019}\cite{proposal-latency-bustos} as a robust and expandable framework for all the algorithms and data structures. Figure \ref{fig:cortex} shows a schematic overview. 

\begin{figure}[h!]
    \centering
    \includegraphics[width=1\linewidth]{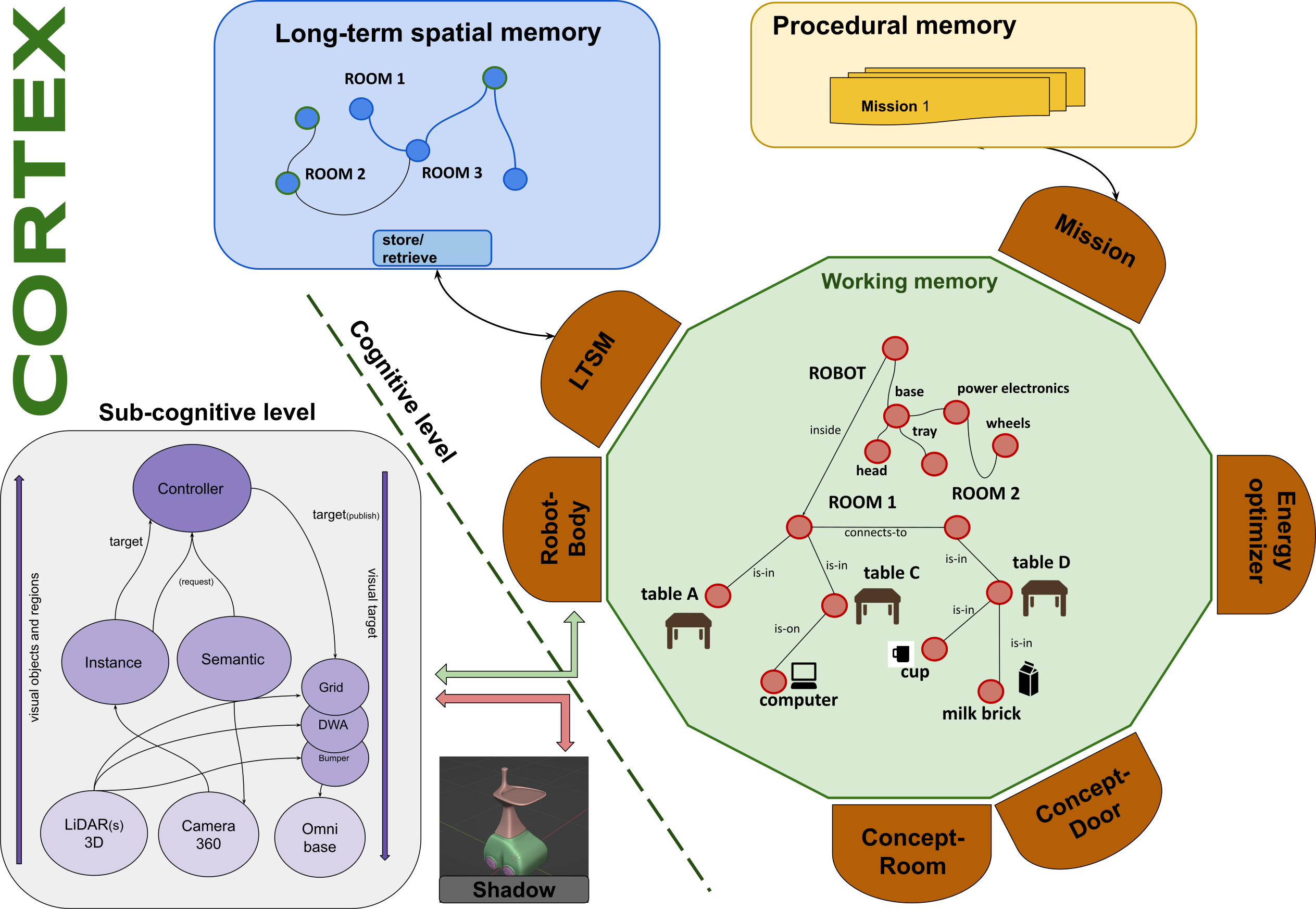}
    \caption{An overview of the CORTEX architecture with the elements used in this work. The sub-cognitive level on the left encompasses low-level perceptive and control components. The working and long-term spatial memories are managed by the six agents depicted as brown shapes. }
    \label{fig:cortex}
\end{figure}

\textcolor{black}{In this paper, we introduce a concept-first architecture where autonomous agents, each representing a spatial concept like room or door, cooperatively and asynchronously build a 3DSG. This approach diverges from the classical metric-first paradigm by allowing the robot to actively choose its next action to refine or expand its representation, building spatial understanding directly from human-interpretable concepts.} \textcolor{black}{We model the world as a set of rectangular rooms connected by doors}. \textcolor{black}{The robot is given a functional definition of the room and door concepts, and the goal is to incrementally build a scene graph representation of the environment by instantiating these concepts and relating their instances geometrically and semantically.} The graph is to be constructed directly from these concepts, avoiding a previous free/occupied grid representation. The robot has a 3D LiDAR that provides a stream of points from the scene. Finally, we use the CORTEX architecture \cite{bustos_cortex_2019}\cite{proposal-latency-bustos} as a robust, expandable framework for all algorithms and data structures. Figure \ref{fig:cortex} \textcolor{black}{depicts} a schematic overview.

% Our first contribution is the idea of a concept process that sustains the life cycle of its instances, created to explain and predict objects of the world. A process for each concept runs asynchronously in the architecture and behaves opportunistically, contributing to building a global, shared representation. In turn, each concept accesses this representation to acquire a necessary context that constrains its creation and update process. 
% Our second contribution is to develop an incremental, scene-graph building system designed for real-time operation in a mobile robot's control architecture as an underlying representational dynamics adaptable to other concurrent tasks. Jointly, the concept processes and the other elements in the architecture collaborate to build and maintain a reliable scene graph that different tasks can use.

\textcolor{black}{Our primary contribution is a novel system of asynchronous concept agents. Each agent sustains the life-cycle of its specific concept, actively creating, managing, and refining instances to explain and predict real-world objects. Each concept agent in the architecture behaves opportunistically, contributing to the construction of a global, shared representation. In turn, each concept \textcolor{black}{agent accesses this shared representation to acquire contextual information} that constrains its creation and update processes according to the hierarchical constraint-propagation principle, in which higher-level concepts guide the detection of lower-level concepts.} 
\textcolor{black}{Our second contribution is a world model composed of connected, local metric frames rather than a single global one. This work is primarily an architectural exploration that demonstrates how this organisation can build persistent, actionable representations directly from concepts.}
\textcolor{black}{Our third} contribution is to develop an incremental, scene-graph building system designed for real-time operation in a mobile robot's control architecture as an underlying representational dynamics adaptable to other concurrent tasks. Jointly, the concept processes and the other elements in the architecture collaborate to build and maintain a reliable scene graph that different tasks can use.

%%%%%%%%%%%%%%%%%%%%%%%%%%%%%%%%%%%%%%%%%%

\section{Related works}

\subsection{From geometric SLAM to semantic and hybrid SLAM}
Classical geometric SLAM has been extensively studied and remains a cornerstone for robot localisation and mapping; however, its lack of high-level semantic content limits human–robot interaction, long-term robustness, and task-oriented reasoning \cite{Cadena2016survey, orb_slam}. To address these shortcomings, numerous works have proposed augmenting metric maps with semantic labels and object-centric landmarks, a family of methods commonly referred to as semantic or metric–semantic SLAM \cite{salas2013slam, bowman2017probabilistic, mccormac2017semanticfusion}. Other approaches tackle persistent mapping by emphasising the need for robust semantic reasoning in dynamic environments \cite{narayana2020lifelongupdatesemanticmaps, santos2017spatio}. Similarly, CNN-based methods such as SemanticFusion \cite{mccormac2017semanticfusion} and Mask R-CNN extensions (e.g., PanopticFusion) \cite{he2017mask, narita2019panopticfusiononlinevolumetricsemantic} have made dense semantic mapping feasible in (near) real-time, bridging perception and mapping pipelines \cite{CNNsurvey}. Surveys and overviews further summarise trends and techniques in this area, covering pipelines that combine detection, data association, and pose/landmark estimation \cite{semanticslam2022, semanticslam2020}. Open-source libraries such as Kimera demonstrate the practical integration of visual–inertial SLAM, dense reconstruction, and semantic labelling to produce metric–semantic maps in real-time \cite{rosinol20203ddynamicscenegraphs}. \textcolor{black}{Although existing methods successfully augment metric maps with object-level semantics, they follow a metric-first paradigm. Our work departs from this by allowing the 3DSG to integrate data based on high-level concepts, which form the primary organisational structure of the representation.}

\subsection{3D Scene Graphs and Dynamic Scene Graphs}
The idea of representing richer semantic and topological structure as a graph grounded in 3D geometry was crystallised by Armeni et al., who introduced the 3D Scene Graph (3DSG) as a unified hierarchical structure linking floors, rooms, objects, and cameras in reconstructed buildings \cite{armeni20193dscenegraphstructure}. Subsequent works have extended and operationalised this representation for robotics. Rosinol et al. introduced \emph{3D Dynamic Scene Graphs} (DSGs), emphasising actionable perception by adding temporal and agent-centric layers and presenting an automated spatial-perception pipeline to produce DSGs from visual–inertial data \cite{rosinol20203ddynamicscenegraphs}. Hydra and related spatial-perception systems build on these ideas and demonstrate real-time 3DSG construction at scale, integrating segmentation, topological reasoning, and loop closure into an online pipeline \cite{hughes2022hydra, hughes2024foundations}. SceneGraphFusion proposed an incremental, learning-based strategy to predict 3DSG from RGB–D sequences using graph neural networks and attention mechanisms suited to partial observations \cite{wu2021scenegraphfusionincremental3dscene}. In parallel, Bavle et al. explored situational graphs as a lightweight, navigation-focused variant of DSGs, highlighting their role in bridging semantic scene understanding and actionable planning \cite{bavle2022situational}. \textcolor{black}{Most state-of-the-art 3DSG systems, despite their rich hierarchical structures, ground their representation in a global metric frame built from a dense reconstruction. Our architectural contribution is a fundamental departure from this approach, as we construct the metric frame based on instantiated spatial concepts themselves.}

\subsection{\textcolor{black}{Extracting high-level spatial structure}}

Within scene graphs, a first level of spatial structuring is provided by the layout of interior buildings, which condense the basic geometric structure of walls, floors, and ceilings. Their integration into a richer topological and semantic representation enables long-term reasoning and planning.
More traditional approaches resort to the use of recognised geometric algorithms like RANSAC on the point cloud generated by SLAM to identify walls \cite{hossein2019building, 8206513, han2021vectorized}, similar to \cite{khanal2024structurepreservingplanarsimplificationindoor}, where the Manhattan world assumption is added. Other works formulate layout reconstruction as an optimisation problem \cite{rs14184675}, resorting to mesh-based normalisation of input point cloud models and then using a constrained optimiser. 
\textcolor{black}{Their main advantage lies in their independence from large data sets and their geometric accuracy in well-structured environments. However, they are sensitive to hyperparameter tuning, occlusions, orthogonality assumptions, and sensor noise.}

A distinct line of research leverages deep neural networks to infer complete room layouts from single panoramic images, with notable examples including LayoutNet \cite{zou2018layoutnet}, FloorNet \cite{liu2018floornet}, and HorizonNet \cite{Sun_2019_CVPR}. These data-driven models exhibit strong generalisation capabilities and can produce geometrically coherent layouts at a low inference cost. However, from a robotics perspective, their utility is constrained by two primary factors. First, they are dependent on large-scale annotated datasets and often presuppose regular, structured environments. More critically, they operate as 'black-box' systems performing a single-shot inference. This monolithic approach offers no mechanism for incremental refinement or reasoning about partial evidence. Consequently, these models struggle with the ambiguity inherent in a robot's typical starting condition—often a position with limited visibility—a scenario where our concept-driven, hypothesis-validation loop is specifically designed to address. \textcolor{black}{While the choice of a specific detection algorithm can significantly impact performance, our work's primary contribution lies in demonstrating a holistic, functional system. We therefore utilized straightforward detection methods to allow the core principles of our architecture to be evaluated independently of their specific perceptual components.}

\subsection{Incremental, opportunistic and active perception} 
While many 3D Scene Graph (3DSG) systems are validated on offline datasets, the robotics community increasingly demands representations that are built incrementally and refined through active perception for online operation. The principle of closing the planning–perception loop is well-established in Active-SLAM, where robot actions are chosen to reduce map uncertainty or gather task-relevant data \cite{burgard1997active, slam_2023}. This is complemented by information-theoretic approaches that provide principled methods for selecting optimal sensing actions under uncertainty \cite{1041446, lluvia2021active}.

More recent work on situational graphs frames scene understanding as the online construction of optimizable, multi-layer representations (metric, topological, semantic) for navigation \cite{bavle2022situational, shaheer2023graph, bavle023fromslamtosituationalawareness, millan2023better}. Collectively, these research efforts motivate architectures where perception is not a passive aggregator of data, but an active process that deliberately seeks information to validate or refute semantic hypotheses—such as the existence of a room or the precise location of a door \cite{slam_2023, BajcsyAT16}.

Our architecture embodies these principles by integrating affordances as context-sensitive controllers within each concept agent. These affordances are pre-activated when their operational preconditions are met. A central \textit{mission-monitoring} agent executes tasks by selectively activating affordances that guide the robot toward its goal, interleaving these actions with others dedicated to constructing and maintaining the internal scene representation. This mechanism provides the foundation for the incremental, opportunistic, and active perception at the core of our system.

\subsection{Open-vocabulary and language-grounded scene graphs}

Recent advances in vision-language models (VLMs) have catalysed a new generation of open-vocabulary 3D Scene Graphs (3DSGs), enabling robots to perceive and reason about an open set of objects and relationships. The dominant pipeline for these systems typically begins with class-agnostic segmentation of RGB-D frames, followed by the extraction and projection of semantic feature vectors into a common 3D frame. The resulting semantically-enriched point cloud is then processed, often with queries to a Large Language Model (LLM), to establish object identities and inter-object relationships.

While these methods demonstrate powerful capabilities for language-grounded querying, our concept-first architecture diverges from this paradigm in several fundamental ways, centred on the principles of incremental construction and the explicit modelling of environmental structure.

A primary distinction is our commitment to incremental, online construction without prior global maps. Systems like HOV-SG \cite{werby24hovsg}, for instance, require a complete metric map of the environment as a prerequisite for extracting floors, rooms, and objects. Navigation is then enabled by a permanent Voronoi graph of the total free space. In contrast, our approach builds its understanding dynamically as the robot explores. Navigation is handled locally on demand within the current room's reference frame, while inter-room pathfinding is a simple graph search in the Long-Term Spatial Memory.   

Furthermore, our architecture prioritises the explicit modelling of spatial structure and connectivity, whereas many open-vocabulary systems are fundamentally object-centric. Works such as Open3DSG \cite{kochopen3dsg} and OVSG \cite{chang2023contextawareentitygroundingopenvocabulary} excel at establishing rich relationships among objects but do not acquire the containing room or the topological connectivity between rooms as first-class entities in the graph. Our work, conversely, treats the "room" concept as the primary semantic and geometric anchor, with the environment's structure serving as the backbone of the scene graph.

This focus enables our core principle of hierarchical constraint propagation, which is architecturally distinct from the bottom-up process common to many VLM-based pipelines. ConceptGraphs \cite{gu2023conceptgraphsopenvocabulary3dscene}, for example, incrementally adds objects but must wait until an entire sequence has been processed to generate object captions and relationships. \textcolor{black}{Spatial relations are first estimated by proximity and then refined by querying an LLM, a process that prevents high-level concepts (such as a room's walls) from actively constraining and guiding lower-level perception (such as finding a door). Our top-down approach is designed around this very principle of semantic constraint.}

Finally, while the field is advancing towards capturing object affordances and functional relationships, as pioneered by OpenFunGraph \cite{Zhang_2025_CVPR}, our work provides the essential spatial-semantic scaffolding for such reasoning. OpenFunGraph masterfully models interactions between objects and their parts, but does not construct a representation of the surrounding spaces or validate that the graph remains anchored to the world as the robot moves. Our architecture is expressly designed to build and maintain this persistent spatial representation, providing the stable, grounded context in which functional and task-level reasoning can reliably occur.

% TODO: REVISAR contribuciones al final de intro
\subsection{Positioning of the present work}
Most existing 3D scene graph construction approaches follow a metric-first paradigm: they build dense occupancy maps or point clouds, then layer semantic information on top through object detection and classification \cite{wu2021scenegraphfusionincremental3dscene}\cite{hughes2022hydra}\cite{rosinol2021kimeraslamspatialperception}. While effective, this approach conceptualises space primarily in terms of free/occupied regions—representations that enable collision-free navigation but provide limited support for high-level reasoning, manipulation planning, or human-robot communication.
We explore an alternative concept-first architecture in which spatial understanding emerges directly from human-interpretable concepts, without requiring prior metric representations. 
Our approach makes four key departures from existing work:
(i) \textit{asynchronous concept agents}, rather than sequential metric-then-semantic processing, we implement concept classes (room and door) as autonomous processes that run asynchronously within the CORTEX cognitive architecture. Each concept agent manages the complete lifecycle of its instances —from initial detection through validation, maintenance, and execution of affordances— competing opportunistically for robot resources,
(ii) \textit{hierarchical constraint propagation}, where room instantiation provides geometric and semantic constraints that significantly improve door detection efficiency and robustness. Unlike unconstrained analysis of occupancy grids, doors are searched within the constrained space of validated room walls, leveraging spatial relationships inherent in indoor environments. 
%(iii) \textit{direct scene graph construction}, allowing the scene graph to be built incrementally from concept instantiations without requiring global metric maps. This enables real-time operation with bounded working memory while maintaining human-interpretable spatial representations suitable for communication and explanation, and
%(iii) \textit{dynamic memory management through long-term spatial memory (LTSM)}, which asynchronously manages the working memory content by removing distant rooms and loading relevant ones based on robot navigation. This maintains only locally accessible spatial representations in active memory, enabling scalable operation in large environments without compromising real-time performance. 
(iii) \textit{world model of local metric frames} where the global representation, managed by the Long-Term Spatial Memory (LTSM), is a topological graph connecting local, concept-centric metric frames. The robot localises itself relative to the current room's frame, not a global world frame. This architectural pattern is a key distinction from systems like Hydra, which, despite their real-time performance, still ground a hierarchical scene graph in a global metric frame. Our contribution lies in exploring this alternative world model structure, which may offer benefits in scalability and human interpretability. 

(iv) \textit{architectural vs. algorithmic contribution}, as this work should be understood as an architectural exploration rather than an algorithmic one. We deliberately use simple, well-understood perception methods (e.g., Hough transforms) to focus attention on the system's organisational principles. The core claim is not that our perception modules are state-of-the-art, but that an architecture of asynchronous, communicating concept agents can effectively build an actionable and persistent scene graph. While we currently focus on predefined concepts, we envision this architecture as a foundation that could be extended to synthesise the code of new concept agents using generative models.

% Taken together, prior work provides a rich foundation for semantically grounded spatial representations. The earliest explicit formalisation of the 3DSG concept is credited to Armeni et al. \cite{armeni20193dscenegraphstructure}, and later contributions (Kimera, DSG, Hydra, SceneGraphFusion, HOV-SG, Open3DSG) progressively addressed real-time construction, dynamics, incremental prediction and open-vocabulary semantics \cite{rosinol20203ddynamicscenegraphs, rosinol2020kimeraopensourcelibraryrealtime, hughes2022hydra, wu2021scenegraphfusionincremental3dscene,  kochopen3dsg}. 
% Our approach departs from many of these works by (i) implementing \emph{concept processes} (e.g., \emph{room} and \emph{door}) that run asynchronously and opportunistically within a distributed cognitive architecture (CORTEX/RoboComp), (ii) building the scene graph incrementally and directly without requiring a dense global occupancy grid, (iii) maintaining a hierarchy of egocentric (robot frame), allocentric (a selectable external object frame, e.g. the current room) and long-term (topological map) representations, (iv) integrating active, self-validating perception loops where concept instances propose, test and refine geometric and semantic hypotheses while competing for robot resources. This combination aims to bring 3DSG construction closer to the operational constraints of real-time robotic control and human–robot interaction.

%%%%%%%%%%%%%%%%%%%%%%%%%%%%%%%%%%%%%%%%%%

\section{Method}

\subsection{Architectural Motivation}
Before describing the CORTEX implementation, we clarify our motivation for concept-first spatial representation. Traditional robotic mapping conceptualises space using occupancy grids—binary free/occupied classifications that enable path planning but provide limited semantic content. While sufficient for navigation, these representations offer little support for: (i) human-robot communication—explaining robot decisions in terms of "occupied cells" rather than "rooms" and "doors"; (ii) high-level task planning requiring reasoning about functional spaces and their relationships; (iii) manipulation constraints—understanding that objects belong to rooms and inherit spatial constraints; and (iv) knowledge transfer—communicating spatial understanding between robots or to humans.

% Concept-first representation addresses these limitations by grounding spatial understanding in human-interpretable entities from the outset. Rather than adding semantic layers post-hoc, the robot builds its world model directly through concept instantiation, enabling transparent reasoning about spatial relationships and their functional implications.
% The key insight is hierarchical constraint propagation: once a room concept is instantiated, it provides geometric boundaries (walls), functional relationships (containment), and semantic priors (door locations) that constrain and improve subsequent concept detection. This differs fundamentally from unconstrained analysis of metric data, where door detection must consider the entire perceptual field rather than focusing on geometrically and semantically relevant regions. This architectural principle naturally extends to future object detection —furniture constrained to floor planes, paintings to walls, or items to table surfaces— progressively narrowing search spaces through accumulated spatial context and reducing computational complexity while improving detection reliability.

Concept-first representation addresses these limitations by grounding spatial understanding in human-interpretable entities from the outset. Rather than adding semantic layers post-hoc, the robot builds its world model directly through concept instantiation, enabling transparent reasoning about spatial relationships and their functional implications.
\textcolor{black}{The central insight of our approach is hierarchical constraint propagation}: once a room concept is instantiated, it provides geometric boundaries (walls), functional relationships (containment), and semantic priors (door locations) that constrain and improve subsequent concept detection. This differs fundamentally from unconstrained analysis of metric data, where door detection must consider the entire perceptual field rather than focusing on geometrically and semantically relevant regions. \textcolor{black}{This architectural principle naturally extends to future object detection — furniture constrained to floor planes, paintings to walls, or items to table surfaces — progressively narrowing search spaces through accumulated spatial context, reducing computational complexity, and improving detection reliability.}

\textcolor{black}{At this stage, the potential efficiency and robustness gains should be understood as intended outcomes of the architectural principle rather than as quantitatively demonstrated results. The idea of reducing the search space through hierarchical constraint propagation follows well-known strategies in perception and planning, and our architecture is designed to incorporate this mechanism in a concept-driven manner. %By presenting the contribution in terms of design intent, we highlight the distinctive value of the concept-first paradigm while leaving room for future empirical evaluation.
}

\subsection{The CORTEX architecture}
 CORTEX is a cognitive robotics architecture initially designed to explore how the robot, the environment, and their interaction can be efficiently represented and anchored. A general scheme is shown in Figure \ref{fig:cortex}. It is organised into two blocks: the cognitive level, a collection of specialised memories interconnected by processes called \textit{agents}, which are responsible for exchanging information among them, and the subcognitive level, which maintains a bidirectional connection with the robot's body. Additional details on the architecture can be found in \cite{proposal-latency-bustos}
 \cite{bustos_cortex_2019}\cite{dsr-icarsc2022}.

The working memory (WM) depicted as a large green circle in Figure \ref{fig:cortex} is a distributed graph data structure implemented to provide very low latency and high throughput to a large set of connected processes that can edit it. The content represents the robot and its environment, which is built and maintained by specialised agents. Nodes in the graph represent objects in the world or the robot itself, and edges represent relationships among them: geometric transformations in SE(3) with uncertainty (called RT edges) or logic predicates. Agents run asynchronously, can edit the graph, and react to changes as part of their local goals. \textcolor{black}{ table~\ref{tab:scenegraph_schema} summarises the types and attributes of elements in the scene graph.}. In the configuration used in this paper, the \textit{robot-body} agent reacts to specific changes in the graph by controlling the robot's movements. The \textit{mission-monitoring} agent updates the graph to ensure the robot's actions are validly serialised. Finally, the \textit{x-concept} agents are pre-defined concept-aware processes that will attach to compatible objects in the environment. They have access to the subcognitive stream of perceptual data and try to \textit{explain out} the parts of it that match their concept class (e.g. room or door in this case) by predicting it in the next step. 

\textcolor{black}{A key feature of the CORTEX architecture is its management of concurrency. The agents operate asynchronously and communicate entirely through edits to the shared Working Memory graph. To prevent race conditions, the WM's underlying implementation ensures that all node and edge modifications are atomic operations. This design choice ensures data integrity without requiring complex and potentially high-latency locking mechanisms. Agents function as reactive processes, monitoring the graph for relevant state changes and posting their own updates in turn. This event-driven, pub-sub-like interaction model, built upon a foundation of atomic edits, is fundamental to ensuring a consistent and coherent state across the distributed system.}

\begin{table}[h]
\centering
\caption{Scene Graph Schema Definition.}
\label{tab:scenegraph_schema}

\begin{tabular}{|p{1.6cm}|p{1.8cm}|p{3cm}|p{6cm}|}
\hline
\textbf{Element} & \textbf{Type} & \textbf{Key Attributes} & \textbf{Description} \\
\hline
\multicolumn{4}{|c|}{\textbf{Nodes: Semantic and Geometric Entities}} \\
\hline
Node & room & ID, geometry, confidence & represents the primary spatial and semantic entity. A room is defined as a container constructed from a set of associated wall and corner nodes. \\
\hline
Node & wall & ID, geometry (plane equation), parent\_room\_ID & A primary structural component of a room. It acts as a geometric reference and a container for door nodes. \\
\hline
Node & corner & ID, geometry (3D point), confidence & A key geometric feature representing the intersection of walls. Corners serve as stable landmarks for robot localisation within a room's reference frame. \\
\hline
Node & door & ID, geometry (pose, dimensions, anchor\_points), confidence, state (open/closed) & represents an articulation point or gateway between spaces. It is geometrically "hung" from a parent wall node. \\
\hline
\multicolumn{4}{|c|}{\textbf{Edges: Relationships between Nodes}} \\
\hline
Edge & RT (Rigid Transform) & SE(3) transform, covariance & Represents a geometric parent-child relationship with uncertainty. Establishes the local reference frames (e.g., room -> wall, wall -> door). \\
\hline
Edge & match & predicate (door1\_ID, door2\_ID) & A semantic link established when the system identifies that two door nodes, instantiated in different rooms, represent the same physical door, enabling loop closure. \\
\hline
Edge & has\_intention & predicate, attributes relating to the status of the mission & A temporary, task-oriented edge originating from the robot agent. It signals a request to perform an action on a target node (e.g., cross a door). \\
\hline
\end{tabular}
\end{table}

\subsection{\textcolor{black}{Agent Interaction and Coordination via the Working Memory}}

\textcolor{black}{
Agent collaboration is achieved implicitly through the shared Working Memory (WM) graph.
\begin{itemize}
    \item \textbf{Communication:} Agents communicate by creating, modifying, or deleting nodes and edges. For instance, the C\_room agent signals its intent to explore by inserting a has\_intention edge into the WM.
    \item \textbf{Coordination:} Agents are data-driven, reacting to specific patterns in the graph. The C\_door agent, for example, remains dormant until a current\_room node is present and validated in the WM. This ensures a natural, hierarchical order of operations (find room, then find doors).
    \item \textbf{Conflict Resolution:} A dedicated mission\_monitoring agent acts as a central arbiter. It observes all has\_intention proposals from other agents. It prioritises them based on the robot's current high-level task, resolving potential conflicts (e.g., choosing to navigate towards a goal rather than further explore a room).
\end{itemize}
}

\subsection{Overview of the spatial representation construction process}

The robot's primary goal is to build an actionable scene graph representation of its environment through incremental concept instantiation and validation. This process involves several asynchronous agents, primarily $\mathcal{C}_{room}$ and $\mathcal{C}_{door}$, that cooperatively construct a shared spatial model within the Working Memory (WM), supported by the \textit{robot\_body}, \textit{energy\_optimiser} and \textit{long\_term\_spatial\_memory} agents, and orchestrated by the \textit{mission\_monitoring} agent.

The representation construction follows a concept-first paradigm, in which spatial understanding emerges from the instantiation and anchoring of semantic entities rather than from metric grid mapping. The robot begins with no prior spatial knowledge and bootstraps its understanding through active exploration. Upon initialisation or after crossing a door, the $\mathcal{C}_{room}$ agent attempts to instantiate a room concept by detecting corners in the 3D LiDAR point cloud $P \in R^3$. Successfully instantiated rooms serve as reference frames for subsequent spatial reasoning, with the robot's pose continuously updated relative to the current room via factor graph optimisation.

\begin{figure}[h!]
    \centering
    \includegraphics[width=0.7\linewidth]{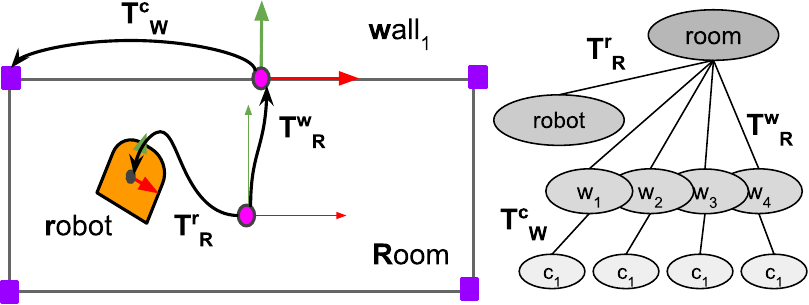}
    \caption{Reference frames used for the elements in the scene graph. On the left, the grey rectangle represents the model room that best fits the corners. $\boldsymbol{T}^{w}_{R}$ denotes the transformation from the room to the wall, so a point $p$ in the wall's frame is transformed to the room's frame as $q = \boldsymbol{T}^{w}_{R}  p$. On the right is the kinematic tree inserted into the WM, where $\boldsymbol{T}^{c}_{W}$ denotes the transformation from the wall to the corresponding right corner in the wall reference system.}
    \label{fig:frames}
\end{figure}

For this initial exploration, we adopt the Manhattan world assumption—rooms are rectangular with orthogonal walls—which simplifies geometric reasoning while remaining applicable to many real-world indoor environments. Each room maintains a centre-based coordinate frame with child frames for its four walls, forming a kinematic tree that structures spatial relationships. Doors must lie within wall boundaries, with at most one door per wall, and each door connects exactly two rooms. \textcolor{black}{Figure \ref{fig:frames} \textcolor{black}{illustrates} the reference systems used in the scene and their hierarchy in graph form.}

\begin{figure}[h!]
    \centering
    \includegraphics[width=1\linewidth]{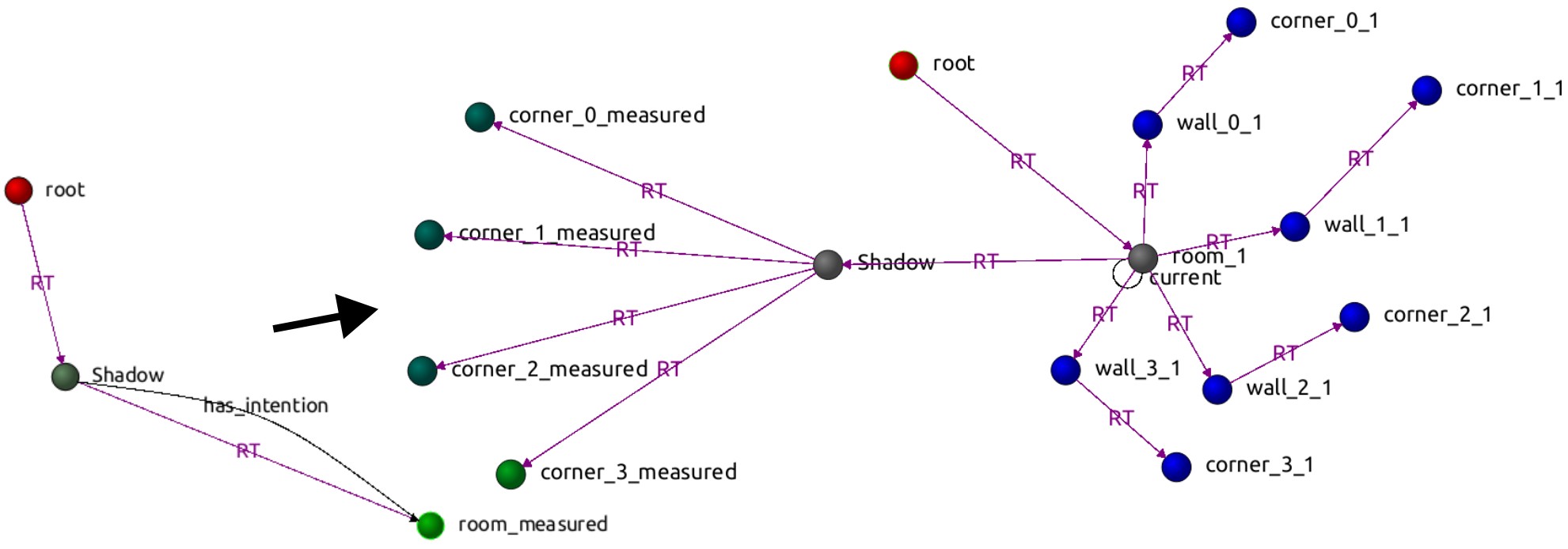}
    \caption{Graph state transition. The robot starts as the graph's initial frame (\textit{root is a dummy node}) and transforms to hang from the new room when it is inserted in the graph. \textit{corner\_x\_measured} are nodes that hold the measurements of the corners matched with the corresponding nominal ones on the right.}
    \label{fig:robot-transition}
\end{figure}

The concept lifecycle encompasses detection, initialisation, validation, and maintenance phases. During initialisation, concept agents propose intentional actions (e.g., navigating toward a room's estimated centre or to a vantage position in front of a door) to gather sufficient observational evidence. These actions are mediated by \textit{has\_intention} edges in the WM and require approval from the \textit{mission\_monitoring} agent before resource allocation. The initialisation process accumulates measurements across multiple robot poses, building temporal pose graphs that resolve geometric ambiguities through optimisation. Once sufficient evidence confirms adherence to concept priors, instances transition from provisional to permanent status in the WM. Figure \ref{fig:robot-transition} shows an example of a structural modification in the scene graph. The transition occurs when a room is detected, initialised and inserted into the graph, causing the robot node to change its parent. The new \textit{RT} edge is now updated by the \textit{energy\_optimizer} agent, minimising the differences between measured and nominal features.

Critical to this architecture is the synchronisation achieved through graph-based communication. The $\mathcal{C}_{door}$ agent remains dormant until a current room exists in the WM, ensuring doors are always grounded within validated spatial contexts. This hierarchical dependency — doors require rooms, rooms enable localisation — creates a natural ordering that prevents premature or unconstrained concept instantiation. Similarly, the \textit{energy\_optimiser} agent continuously monitors the WM for validated corner and door measurements, incorporating them as landmarks in its localisation factor graph only after concept agents confirm their validity through successful matching against predictions.

% The architecture explicitly acknowledges operational boundaries. Severely cluttered environments where furniture occludes wall geometry, non-rectangular spaces that violate Manhattan assumptions, or dynamic scenes with moving obstacles may challenge successful concept instantiation. Rather than attempting exhaustive coverage of edge cases, the system maintains probabilistic beliefs and can defer instantiation when evidence remains insufficient. Despite this cautious approach, the current implementation faces two critical failure modes: false positives, where incorrect models satisfy the validation criteria and become accepted as valid representations, and model invalidation, where initially correct models become inconsistent as additional exploration reveals previously unobserved geometric features. We are developing backtracking mechanisms to revert to previous stable states when model confidence degrades, as well as dynamic restructuring capabilities to adapt existing representations when new evidence contradicts established beliefs. This work is discussed further in the \textit{Conclusions and future works} section. The \textit{mission\_monitoring} agent can interrupt ongoing initialisation processes to pursue higher-priority goals, leaving partial representations for later refinement. This opportunistic approach allows the robot to build useful, if incomplete, spatial models while remaining responsive to task demands.

Through this distributed, asynchronous process, the robot incrementally constructs a scene graph that directly supports navigation, manipulation planning, and human communication without requiring intermediate metric representations. The resulting graph maintains both geometric precision (through continuous localisation) and semantic richness (through concept instantiation), providing an actionable spatial model grounded in human-interpretable entities.

\subsubsection{\textcolor{black}{Assumptions and limitations}}

\textcolor{black}{The architecture explicitly acknowledges operational boundaries. Severely cluttered environments where furniture occludes wall geometry, non-rectangular spaces that violate Manhattan assumptions, or dynamic scenes with moving obstacles may make successful concept instantiation difficult. Rather than attempting exhaustive coverage of edge cases, the system maintains probabilistic beliefs and can defer instantiation when evidence remains insufficient. \textcolor{black}{Despite this cautious approach, the current implementation faces two critical failure modes: false positives, in which incorrect models satisfy the validation criteria and are accepted as valid representations, and model invalidation, in which initially correct models become inconsistent as additional exploration reveals previously unobserved geometric features.} We are developing backtracking mechanisms to revert to previous stable states when model confidence degrades, as well as dynamic restructuring capabilities to adapt existing representations when new evidence contradicts established beliefs. This work is discussed further in the \textit{Conclusions and future works} section. The \textit{mission\_monitoring} agent can interrupt ongoing initialisation processes to pursue higher-priority goals, leaving partial representations for later refinement. This opportunistic approach allows the robot to build useful, if incomplete, spatial models while remaining responsive to task demands.}

\subsection{Conceptual agents and intentional actions} %room and door

% \par
A conceptual agent \( \mathcal{A}_c \) is an autonomous process responsible for the instantiation, maintenance, and refinement of a specific concept class \( \mathcal{C}_c \in \mathbb{C} \), where \( \mathbb{C} \) denotes the set of all concept classes manageable by the robot (e.g., \( \mathcal{C}_{\text{room}}, \mathcal{C}_{\text{door}}\)). Each concept class \( \mathcal{C}_c  \) is defined as a tuple $\mathcal{C}_c  = \left( \mathcal{F}, \Phi, \mathcal{A}_{\text{ff}}, \mathcal{I}, \mathcal{L} \right)$, where:

\begin{itemize}
  \item \( \mathcal{F} \) is the set of $f_i$ measurable properties that define the geometry of the concept (e.g., height, width, centre, corners, anchor points, etc.).
  \item \( \Phi \) is a set of prior values over the properties that determine when a new instance is created. (e.g., height to width ratio).
  \item $\mathcal{A}_{\text{ff}}$ is the set of affordances associated with the concept. In this case, the affordances are \textit{visit} for the room and \textit{visit} and \textit{cross} for the door. The \textit{visit} intention is notified to the \textit{mission\_monitoring} agent with the insertion of a \textit{has\_intention} edge in the graph, while the \textit{cross} affordance is notified with the creation of a special node \textit{aff\_cross\_x} \textcolor{black}{(denoting an affordance to cross door x)} hanging from the associated door. If the affordance is accepted by the \textit{mission\_monitoring} agent, the \textit{robot\_body} agent executes the action until completion.
  \item $\mathcal{I}$ is the initialisation process for a new instance candidate. 
  \item \(  \mathcal{L}\) is the life cycle of the concept instances, defined as a behaviour tree that monitors and controls the transitions between internal states.
\end{itemize}

\begin{figure}[h!]
    \centering
    \includegraphics[width=1 \linewidth]{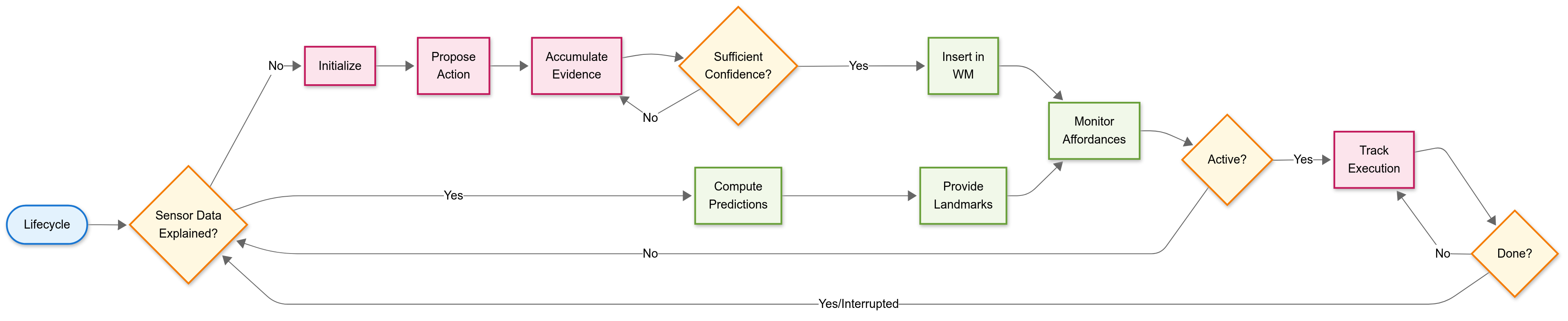}
    \caption{Flow chart showing the life-cycle of a concept. The two outgoing lines of the first decision box cover the insert new instance and update existing instances situations. }
    \label{fig:life-cycle}
\end{figure}

The lifecycle ( $\mathcal{L}$ ) is implemented as a behaviour tree that continuously orchestrates four primary stages of concept instance management, see Figure \ref{fig:life-cycle}. First, the agent evaluates whether existing nominal instances in the WM can explain incoming sensor measurements through forward prediction and matching. When sensor data cannot be adequately explained by current instances (matching error exceeds threshold), the initialisation process ( $\mathcal{I}$ ) is triggered, accumulating evidence through intentional actions until sufficient confidence permits instantiation of a new nominal element in the WM. For established instances, the agent maintains their validity by computing predicted feature positions based on current robot pose estimates, enabling these predictions to serve as landmarks for robot localisation through the energy minimisation process. Concurrently, when affordances are active, the agent monitors their execution status through the robot node, tracking action progress until completion or interruption by the mission monitoring agent. \textcolor{black}{This cyclic process ensures that concept instances remain grounded in sensory evidence, provides stable reference frames for navigation, and supports opportunistic exploration through affordance execution.}

% \begin{figure}
%     \centering
%     \includegraphics[width=1 \linewidth, height=5cm]{figures/life-cycle.png}
%     \caption{Flow chart showing the life-cycle of a concept. The two outgoing lines of the first decision box cover the insert new instance and update existing instances situations. }
%     \label{fig:life-cycle}
% \end{figure}

\begin{figure}[h!]
    \centering
    \includegraphics[width=1 \linewidth]{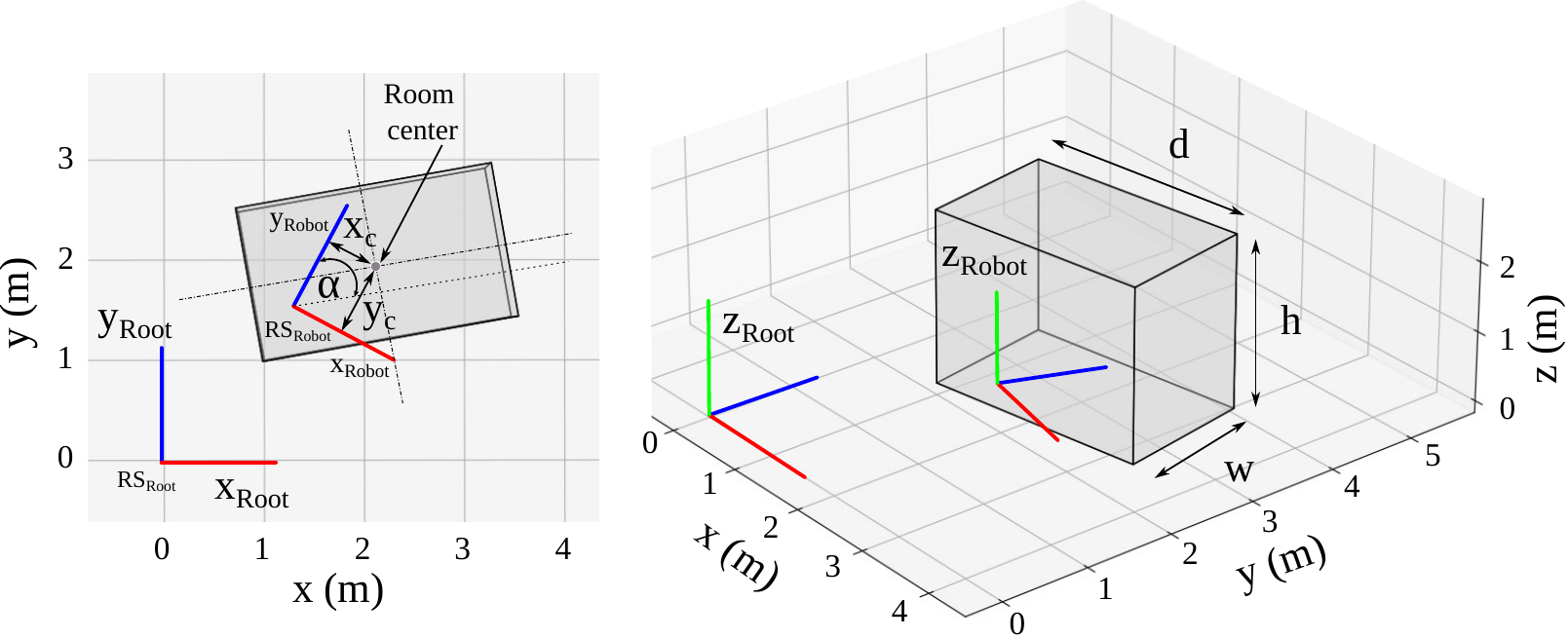}
    \caption{$\mathcal{F}_{room}$ components diagram. Before the room is established, two reference systems are established: the origin reference system and the robot reference system. In the case of the robot, $x_{c}$ and $y_{c}$ denote the position of the centre of the room relative to the robot, while $\alpha$ is the angle relative to the longest side. Conversely, $w$ denotes the length of the shorter side, $d$ represents the length of the longer side, and $h$ signifies the height of the room.}
    \label{fig:rs_figure}
\end{figure}

\subsubsection{\textit{Room-Concept} Agent}

\( \mathcal{C}_{room} \) implements the room concept class. The class provides a parameterisation of a room in terms of centre, angle and dimensions, $\mathcal{F}_{room} = [\textcolor{black}{x_{c}},\textcolor{black}{y_{c}},\alpha,w,d,h]$, as well as the equivalent description by its corners $c_i \in C$. \textcolor{black}{$\mathcal{F}_{room}$ components are illustrated on Figure \ref{fig:rs_figure}}. Room instances are actively searched when the robot starts or when it crosses a door. \textcolor{black}{As explained in the next subsection, the \textit{long-term spatial memory} agent asynchronously recovers previously known rooms and inserts them into the WM.}

Room corners are detected using a simple yet efficient method described in Algorithm~\ref{alg:room_detector}. This algorithm, along with the door detection method presented later, does not represent state-of-the-art (SOTA) performance but provides a sufficient baseline for demonstrating the proposed architectural principles.

\begin{algorithm}[h!]
\caption{Corners Detector Algorithm}
\begin{algorithmic}[1]
    \State $p_i=P(\rho, \phi, k),  z_{min} <  p_z < z_{max}$
    \State $\mu = \frac{1}{M}\sum{p_i}$
    \State $\Sigma = Cov(p)$
    \State $\overline{c} = \mu$
    \State $\overline{s}=\lambda_1(\Sigma), \lambda_2(\Sigma)$
    \State $l=Hough(p)$
    \State $l\in L^* if |l| > L_{min}$
    \State $\hat{c}=\begin{pmatrix} |l| \\ 2 \end{pmatrix}$
    \State $\hat{c}\in \mathbf{C} \iff \textcolor{black}{l_i\langle l_j}=\frac{\pi}{2}$
    \State Return $\mathbf{C}$
\end{algorithmic}
\label{alg:room_detector}
\end{algorithm}

The algorithm begins by (1) filtering the 3D LiDAR field $P$ to retain only those points \(p_i\) within a vertical slice defined by \(z_{\min} < p_z < z_{\max}\) reducing noise from floor and ceiling artifacts; (2-5) the mean and covariance matrix are computed to estimate the room centre \(\overline{c}\) and room's dimensions $\overline{s}$; (6) a Hough Transform is applied, \textcolor{black}{using OpenCV’s point set implementation to directly process the point cloud without rasterization}, in order to extract structural lines; (7) candidate walls \(L^*\) are retained if their lengths exceed a threshold $L_{\text{min}}\textcolor{black}{=1m}$; (8,9) all pairs of lines \textcolor{black}{($l_i, l_j$)} are checked for $\pi/2$ angle crossings to select candidate corners \(\hat{c}\). The algorithm returns a list of corners. \\

% \begin{algorithm}[h]
% \caption{Corners Detector Algorithm}
% \begin{algorithmic}[1]
%     \State $p_i=P(\rho, \phi, k),  z_{min} <  p_z < z_{max}$
%     \State $\mu = \frac{1}{M}\sum{p_i}$
%     \State $\Sigma = Cov(p)$
%     \State $\overline{c} = \mu$
%     \State $\overline{s}=\lambda_1(\Sigma), \lambda_2(\Sigma)$
%     \State $l=Hough(p)$
%     \State $l\in L^* if |l| > L_{min}$
%     \State $\hat{c}=\begin{pmatrix} |l| \\ 2 \end{pmatrix}$
%     \State $\hat{c}\in \mathbf{C} \iff \hat{c_1}\langle\hat{c_2}=\frac{\pi}{2}$
%     \State Return $\mathbf{C}$
% \end{algorithmic}
% \label{alg:room_detector}
% \end{algorithm}

The $\mathcal{C}_{room}$ agent continuously runs the corner detection algorithm.  When a set of detected corners satisfies the instantiation conditions \( \Phi_{room} \), and there is no room object in the WM already, the $\mathcal{C}_{room}$ agent inserts a temporary room node into the WM, linked to the robot node via a \textit{has\_intention} relation. Upon approval by the \textit{mission-monitoring} agent, computational and physical resources are allocated, and the robot initiates a navigation action toward the estimated centre of the room, computed as the centroid of the currently measured corner points. This action develops locally in the robot's frame.

At the beginning of the action, the robot's position is taken as the initial zero frame. \textcolor{black}{As the action proceeds, new poses are concatenated from the robot's odometry readings and stored in a local buffer}. The measured corners are also stored in the current robot frame.
When the action is completed, the concept agent evaluates whether the accumulated data is sufficient; if so, it initiates a post-processing phase. Otherwise, sampling continues from the new vantage point. The \textit{mission-monitoring} agent can revoke the affordance at any time.

\textcolor{black}{As no external frame is available, the agent builds a temporal pose graph in the GTSAM\footnote{https://github.com/borglab/gtsam} framework using data accumulated during the action. The measured corners serve as landmarks, and because they are observed from multiple robot poses, the graph acquires the necessary redundancy. The optimal graph that minimises the sum of prediction errors provides a set of corrected landmarks and poses. From the set of corrected corner poses, a histogram is built to select the most probable corners that comply with the room constraints. The concept agent applies additional priors at this point to potential candidates, discarding rooms that are too large or too small. }

The process ends with inserting the room into WM, along with its walls and corners, and an \textit{RT} edge that anchors the robot to the room. This last step requires some elaboration, since a rectangle lacks an intrinsic orientation. The temporal pose graph was anchored to the action start pose. The selected set of corners has coordinates in that frame, and the robot now occupies the last pose of the graph. Using these data, we want to set the room frame as the initial frame and compute the robot's pose with respect to it. \textcolor{black}{A simple solution is to take the position and size room parameters from the selected corners, and compute the rotation, or equivalently, an enumeration of the corners that assigns index zero to the first corner to the left of the robot heading direction.}

\subsubsection{\textit{Door-Concept} Agent}

The $\mathcal{C}_{door}$ agent controls the life cycle of instances belonging to the door concept class. Doors are parametrised in $\mathcal{F}_{door}$ as pairs of anchor points defining the door ends. They have two affordances $\mathcal{A}_{ff}$: \textit{visit} and \textit{cross}. \( \Phi_{door} \) is the range of admissible distances between the anchor points, currently set to $0.5 - 1.3$ meters.

\par

We propose a simple door detection algorithm, shown in Algorithm~\ref{alg:door_detector}.  It begins by (1) assuming that doors lie within the walls $W$ of the room and filtering out any point \( p_i \) whose distance to the walls exceeds a predefined threshold $\delta=0.15m$; (2) extracts a polar 2D scan \( \rho = p_i(\phi, k = K) \), where \( \rho \) is the measured distance for each angular value \( \phi \) at a fixed height index \( K = 1.7m \), in our experiments; (3) takes the derivative \( \dot{\rho} = \frac{\partial \rho} {\partial \phi} \) to identify abrupt changes in the distance dimension; (4) applies the Heaviside step function \( H \) to yield a binary peak map where a value of 1 indicates a potential edge and zero otherwise. \textcolor{black}{$\lambda=0.5m$ value is a distance threshold between continuous points}; (5) computes all combinations of pairs of detected points; (6) each candidate is tested against the width constraint and the wall constraint that ensures that the line segment \textcolor{black}{which is defined between detected edge points $p$ and $q$}, lies within the set of wall points; (7) returns the set \( D \) of valid detected doors.

\begin{algorithm}[h!]
\caption{Door detector algorithm}
\begin{algorithmic}[1]

    \State $p_i = p \in P \wedge  \mathrm{dist}(p, W) \leq \delta $  
    % \STATE $\rho = L^{3D}(\phi, k = K)$
    \State $\rho = p_i(\phi, k = K)$
    \State $\dot{\rho} = \frac{\partial \rho}{\partial \phi}$
    
    \State $P = H(\dot{\rho} - \lambda) = 
    \begin{cases} 
    0 & \text{if } \dot{\rho} < \lambda \\
    1 & \text{if } \dot{\rho} \geq \lambda 
    \end{cases}$
    
    \State $\left\{\hat{d}_{i,j}\right\}^N = \begin{pmatrix} |\mathbf{P}| \\ 2\end{pmatrix}$
    
    \State $\hat{d}_{i,j} \in \{D\} \text{ if } 
    \begin{cases} 
    D_{\min} < \|\hat{d}_{i,j}\| < D_{\max} \\
    d_{i,j} = p + t(q - p), \quad 0 \leq t \leq 1, \vec{pq} \in W
    \end{cases}$
    
    \State Return $D$
\end{algorithmic}
\label{alg:door_detector}
\end{algorithm}

% \begin{algorithm}[h]\small
% \caption{Door detector algorithm}
% \begin{algorithmic}[1]
%     \STATE $p_i = p \in P \wedge  \mathrm{dist}(p, W) \leq \delta $  
%     % \STATE $\rho = L^{3D}(\phi, k = K)$
%     \STATE $\rho = p_i(\phi, k = K)$
%     \STATE $\dot{\rho} = \frac{\partial \rho}{\partial \phi}$
    
%     \STATE $P = H(\dot{\rho} - \lambda) = 
%     \begin{cases} 
%     0 & \text{if } \dot{\rho} < \lambda \\
%     1 & \text{if } \dot{\rho} \geq \lambda 
%     \end{cases}$
    
%     \STATE $\left\{\hat{d}_{i,j}\right\}^N = \begin{pmatrix} |\mathbf{P}| \\ 2\end{pmatrix}$
    
%     \STATE $\hat{d}_{i,j} \in \{D\} \text{ if } 
%     \begin{cases} 
%     D_{\min} < \|\hat{d}_{i,j}\| < D_{\max} \\
%     d_{i,j} = p + t(q - p), \quad 0 \leq t \leq 1, \vec{pq} \in W
%     \end{cases}$
    
%     \RETURN $D$
% \end{algorithmic}
% \label{alg:door_detector}
% \end{algorithm}
% \vspace{-2mm}

\begin{figure}[h!]
    \centering
    \includegraphics[width=1\linewidth]    
    {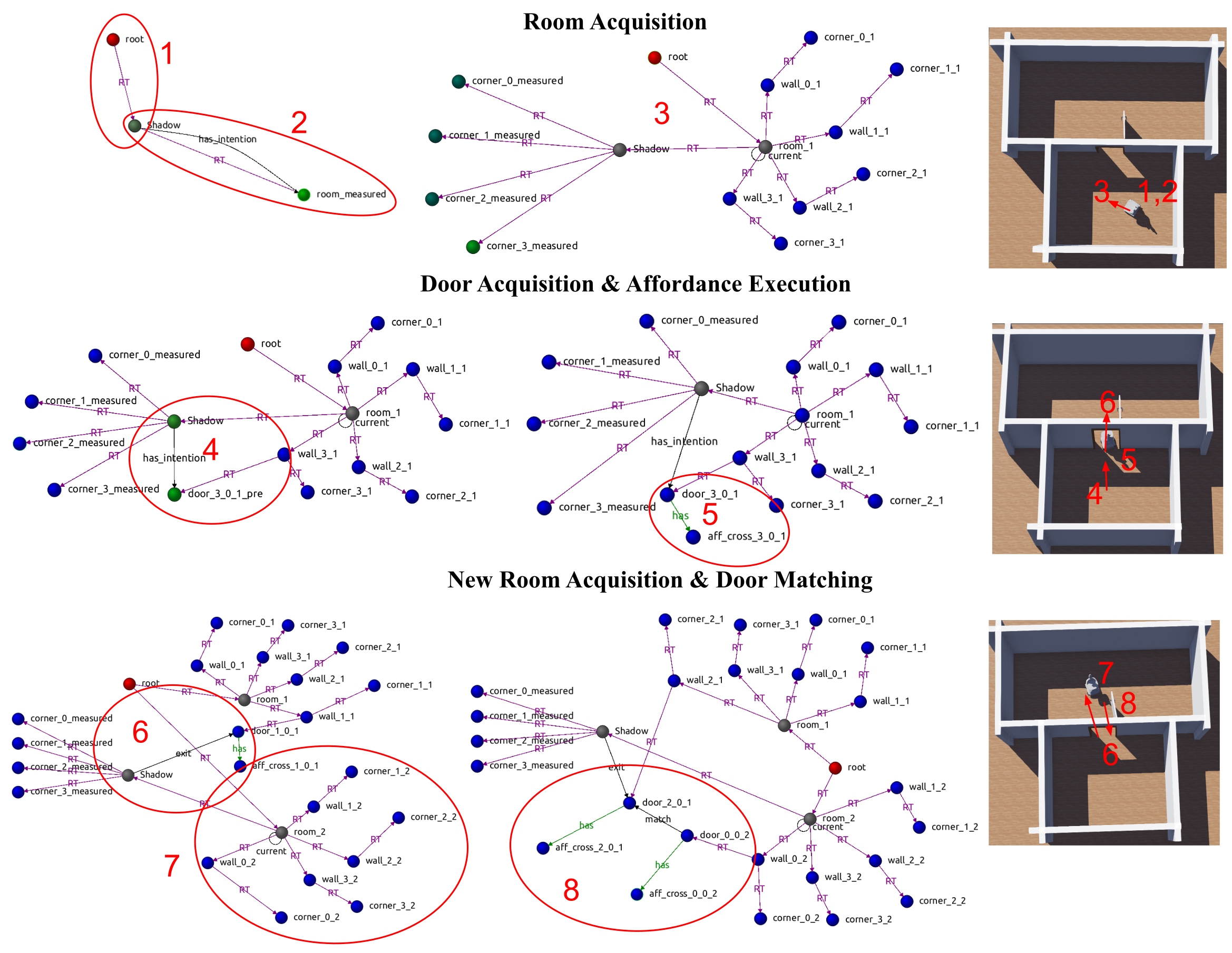}
    \caption{Main stages in constructing the scene graph during exploration, reflecting transitions as different concept instances are initialised and affordances are executed. See text for a detailed explanation.}
    \label{fig:all_states}
\end{figure}

When the current room is detected in the WM, the $\mathcal{C}_{door}$ agent starts running the door detection algorithm. It silently returns if existing nominal doors can explain all the pairs of anchor points detected in the point cloud. When a new door is detected, a process similar to the one described before starts, but with some differences. A provisional door, labelled as \textit{door\_x\_pre} \textcolor{black}{to signify its preliminary, unvalidated status as a concept instance}, is inserted into WM hanging from its containing wall, and a \textit{has\_intention} edge is inserted between the robot and the door. Figure \ref{fig:all_states} shows how these changes occur in the scene graph.

Once the \textit{mission\_monitoring} agent assigns resources, the \textit{visit} affordance is executed by the \textit{robot\_body} agent. The robot navigates towards a point located $1m$ from the door's centre. But now the robot is being localised in the room's frame by the \textit{energy\_optimiser} agent running a global optimisation over the scene graph. To benefit from this situation, the $\mathcal{C}_{door}$ agent only updates the matching between the measurements and the provisional anchor points and waits for the end of the action. The \textit{energy\_optimiser} agent recognises those objects as landmarks and includes them in the minimisation. When the action finishes, the fine-tuned anchor points are set as permanent, and their nominal coordinates are computed and inserted in the \textit{RT} edge connecting the wall to the new door. The door name is changed to \textit{door\_x}, removing the \textit{pre} suffix. \textcolor{black}{The nominal door now provides two additional fixed landmarks for the \textit{energy\_optimiser} agent to update the robot's pose in the room.} Finally, a node connected to the door with an \textit{has} edge is added to show that a \textit{cross} affordance is available for the \textit{mission\_monitor} agent to include in its planning logic.

\subsection{Long-Term Spatial Memory}

\textcolor{black}{The proposed architecture deliberately sacrifices global metric consistency in favour of local precision and long-term scalability. This is managed by the CORTEX long-term spatial memory (LTSM) agent, whose primary function is to maintain a topological graph of locally consistent, concept-centric metric frames.} The LTSM stores previously visited rooms and their connections via doors. This information is maintained in a persistent graph data structure using the \textit{igraph} library\footnote{https://igraph.org/}. The agent operates asynchronously over the WM, removing visited rooms when the robot has entered and initialised a new one, and bringing (i.e., predicting) known rooms back when the robot is about to enter a previously explored one. The precise timing of these actions is explained below.

When the robot enters an unfamiliar room, it needs to retain some information about the room it is leaving until a reliable connection can be made between the two. The agent stores doors as dual objects with coordinates in both rooms' frames. This is the minimal information needed to maintain an actionable graph of connected rooms. As the agent observes the crossing, it waits until the new room is initialised following the procedure described before, and only then sets the new room as the current room and removes the old one from the WM. The required synchronisation is done through the edition of node attributes and edges in the WM, since this is the only way for agents to communicate. 

\textcolor{black}{The insertion of a known room is simpler since this agent is anticipating what $\mathcal{C}_{room}$ will see. The critical variable is the orientation of the new room relative to the robot, since the success of the corner-matching process, which determines acceptance, depends on it. The information is taken from the door dual coordinates. The new room is oriented by first transforming its corners to the door's coordinate frame.} 
Then, we identify both door frames as being at the same point in space, and use this to transform the corners back to the old room's coordinate frame and, finally, to the robot's frame. In this way, the robot can match the new room's nominal corners with fresh measurements since both are in the same frame, and confirm the recognition of the new room. Hereafter, the old room is removed, and the robot is changed in the scene graph to hang now from the new room. 

\par

\begin{figure}[h!]
    \centering
    \includegraphics[width=0.7\linewidth]{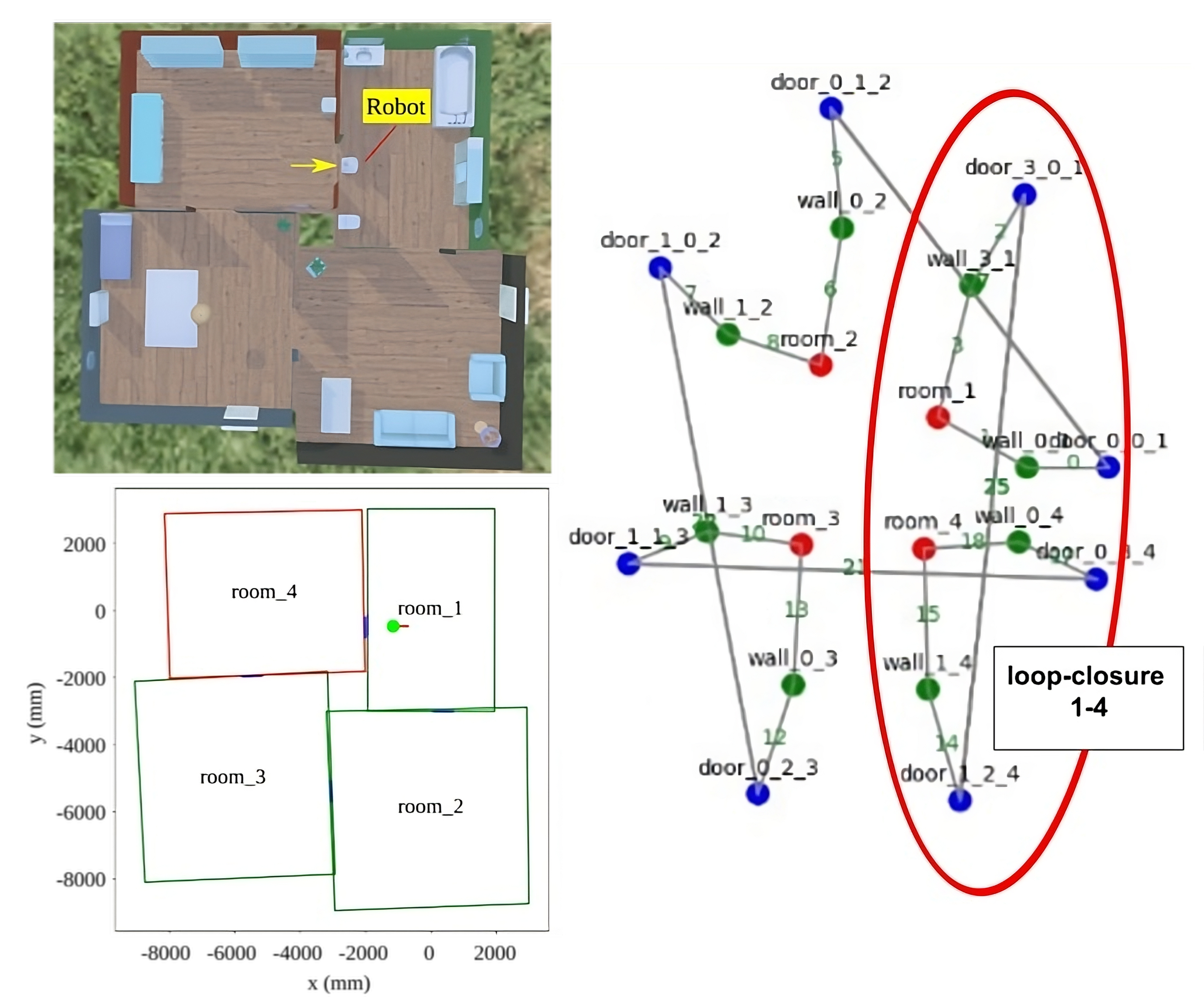}
    \caption{LTSM room representation \textcolor{black}{during a transition from a known room to another known room via an uncrossed door}. The scenario used in the Webots simulator for the experiments is shown in the upper-left image. The down-right area shows a metric reconstruction of the graph on the right. The right image shows the network generated and stored by the \textit{LTSM} agent after the robot has toured rooms 1 to 4. The red circle marks the loop closure between rooms 4 and 1 as an edge connecting door\_3\_0\_1 and door\_1\_2\_4.}
    \label{fig:metric_rooms}
\end{figure}

Figure \ref{fig:metric_rooms} shows a metric representation of the rooms in the four-room scenario, with the rooms the robot has previously traversed shown in green, and the room it is currently located in shown in red. \textcolor{black}{Because the robot is currently executing the cross affordance during the capture, a momentary state lag occurs: the system has not yet loaded the map data for the destination room and therefore remains localized within the room of origin}. Additionally, a green dot indicates the target position the robot is heading towards after crossing the door. The agent can create a metric representation of the graph at any time, as shown in the bottom-left image. \textcolor{black}{The evident alignment errors are due to noise injected into the synthetic LiDAR sensor and the robot's displacements, and could be corrected by constrained optimisation of the rooms' centres, sizes, and orientations, with the doors serving as fixed, shared points between them. It is left for future work, as it does not interfere with the goals of this paper. The red circle on the right shows the result of a loop closure between rooms 1 and 4 through their shared door. Currently, the \textit{LTSM} agent computes a loop closure by projecting the robot's pose into all rooms as it approaches a new one. A more robust method based on visual descriptors and the objects stored in the room is under construction. }

\subsubsection {Scalability Through Hierarchical Memory Organisation}
The LTSM component addresses a fundamental scalability challenge in robotic spatial representation: maintaining complete world models in active memory becomes computationally prohibitive as environment size increases. The CORTEX architecture's memory organisation provides several mechanisms that enable operation in arbitrarily large environments while preserving real-time performance. First, the architecture employs dynamic memory management based on the principle that robots require detailed spatial representations only for their immediate operational context\textcolor{black}{, thereby avoiding the computational burden of maintaining and continuously optimising a single, monolithic global map}. As the robot navigates multi-room environments, the LTSM dynamically loads rooms into working memory based on current location and planned actions, while unloading distant or irrelevant spaces. This selective memory management maintains bounded computational requirements regardless of total environment size, enabling real-time operation in large-scale facilities such as office buildings, hospitals, or multi-floor residential complexes.
Second, the LTSM maintains persistent topological graphs representing complete known environment structures, stored using the \textit{igraph} library. These graphs naturally accommodate arbitrary environment topologies—from linear corridor sequences to complex branching structures with multiple floors, wings, and interconnected spaces.
Third, the memory architecture supports hierarchical spatial organisation, with concepts extending beyond individual rooms. Spaces can be organised into floors, buildings, or functional regions (e.g., "residential wing", "laboratory area"). This hierarchical organisation enables efficient navigation planning across multiple scales while maintaining the human-interpretable structure that motivates the concept-first approach. High-level spatial queries can be resolved at appropriate abstraction levels without requiring detailed geometric computation.
And fourth, context-dependent predictive loading activates when the robot approaches transition points (doors leading to known rooms) and anticipates spatial context requirements by pre-loading relevant representations. This includes not only the target room's geometry and contained objects, but also connected spaces that might become relevant for navigation alternatives or task planning. This predictive loading strategy ensures smooth spatial transitions while minimising working memory overhead and maintaining responsiveness.

\textcolor{black}{Taken together, these mechanisms ensure that the system's computational complexity remains bounded and does not grow with the size of the total explored environment. A formal asymptotic analysis is outside the scope of this architectural work; however, a qualitative analysis indicates that high-frequency, computationally intensive operations (e.g., factor graph optimisation, point cloud processing) are constrained to entities within the currently active room in Working Memory. By dynamically loading and unloading spatial contexts, the LTSM prevents the unbounded growth of active nodes and factors in real-time processes. \textcolor{black}{Consequently, the architecture's computational load is not a function of the global environment's size but instead depends on the local complexity of the robot's immediate surroundings.} This design choice is fundamental to ensuring long-term operational performance and scalability, while preserving the transparency and explainability advantages of the concept-driven approach.}

\subsection{Navigating through the scene graph}

Once a nominal current room is established, the \textit{energy-optimiser} agent updates the robot’s position at a fixed frequency of 10 Hz. This agent maintains a GTSAM factor graph. It is initialised with the nominal room corners and the robot’s pose in the WM. As fresh odometry data arrives from the robot sensors, the agent inserts new robot poses into the graph up to 30 nodes, applying a FIFO policy. \textcolor{black}{The sliding-window version of iSAM2 in GTSAM computes the optimal robot pose after the graph modifications, accounting for the required changes to the covariance matrix.} 

\par

% At each insertion of a new robot pose, the \textit{energy-optimiser} agent verifies if there are measured corner values in the WM that the corresponding concept agent has validated. Validation results from a matching process between the projection of the nominal corner on the robot's frame, and the current measurement. We use the Mahalanobis distance to match the corners weighted by the covariances of the measurement and the uncertainty contributed by the current robot pose. The validation is determined by inspecting a Boolean attribute stored in the corner measurement node within the WM. Figure \ref{fig:optimiser} shows how all these elements are related as a factor graph.

\begin{figure}[h!]
    \centering
    \includegraphics[width=0.7\linewidth]{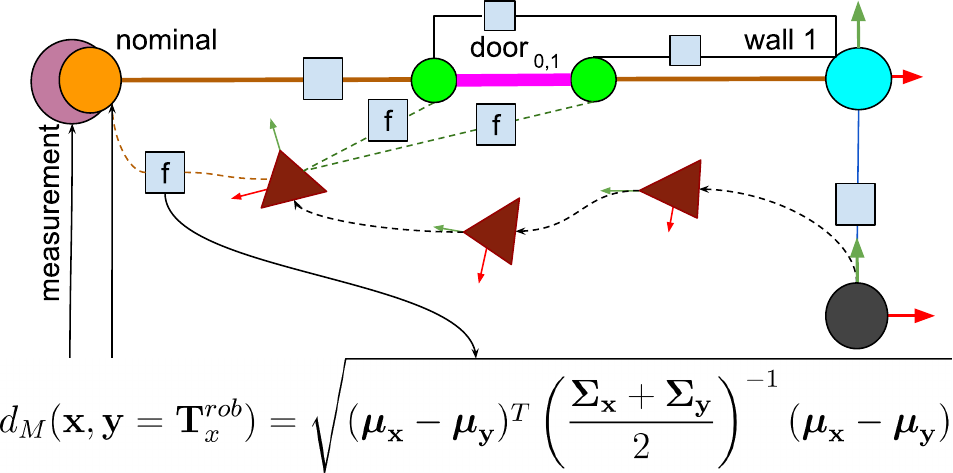}
    \caption{Schematic view of a robot displacement starting at the black circle and advancing through the upper left quadrant. Small grey squares with an \textit{f} are factors, i.e., positive functions measuring the difference between two variables, that link the robot to the objects' parts, corners, and anchor points. Plain grey squares are fixed constraints between elements. The sum of all differences accounts for the system's free energy and is absorbed by the set of past robot poses depicted as red triangles. As more objects are modelled in the WM, more factors are created, making the robot's localisation more robust.}
    \label{fig:optimiser}
\end{figure}

At each insertion of a new robot pose, the \textit{energy-optimiser} agent verifies if there are measured corner values in the WM that the corresponding concept agent has validated. \textcolor{black}{This validation results from a matching process between the projection of the nominal corner $\mathbf{y}$ onto the robot's frame and the current measurement $\mathbf{x}$. We use the Mahalanobis distance $d_M$ to perform this critical pre-association check against a statistical threshold, where $d_M$ is calculated as:
$$d_{M}(\mathbf{x}, \mathbf{y}) = \sqrt{(\boldsymbol{\mu}_{\mathbf{x}} - \boldsymbol{\mu}_{\mathbf{y}})^T \left(\frac{\mathbf{\Sigma}_{\mathbf{x}} + \mathbf{\Sigma}_{\mathbf{y}}}{2}\right)^{-1} (\boldsymbol{\mu}_{\mathbf{x}} - \boldsymbol{\mu}_{\mathbf{y}})}$$
If the match is validated, a factor (cost function) is created in the graph. As illustrated in Figure \ref{fig:optimiser}, this factor (represented by $\textit{f}$) formalizes the statistically weighted difference between $\mathbf{x}$ and $\mathbf{y}$ $(\boldsymbol{\mu}_{\mathbf{x}} - \boldsymbol{\mu}_{\mathbf{y}})$. In this metric, $\mathbf{\Sigma}_{\mathbf{x}}$ represents the covariance of the measurement, and $\mathbf{\Sigma}_{\mathbf{y}}$ represents the total propagated uncertainty of the predicted feature, combining the uncertainty of the nominal corner's position and the robot's current pose}. \textcolor{black}{Validation is determined by inspecting a Boolean attribute stored in the corner measurement node of the WM.}

\textcolor{black}{If a corner has been validated, it is incorporated into the factor graph as a measurement associated with its corresponding landmark. This measurement is connected to the pose node whose timestamp most closely matches the corner observation's timestamp. Similarly, if a door has been instantiated in the WM, its nominal position is inserted into the graph as a landmark. Door measurements are handled in the same way as corner observations.}

\section{Results}

\textcolor{black}{We have designed a series of preliminary experiments} to evaluate the architecture using the Webots simulator. The scenarios include a digital replica of our \textbf{Shadow} mobile robot \cite{Torrejón2024}, with realistic white noise added to the synthetic LiDAR and simulated delays in command execution. \textcolor{black}{All detection parameters and thresholds used in the following experiments, such as those in Algorithm \ref{alg:room_detector} and  \ref{alg:door_detector}, were tuned empirically to suit this specific robotic setup and the characteristics of our test environments.} 

\textcolor{black}{It is worth noting that these experimental scenarios are simplified, structured environments primarily intended as a proof-of-concept to demonstrate the architecture's core mechanics.} Within these constraints, we included rooms of varying dimensions and with different numbers of doors to explore the system's ability to handle some topological variety, while maintaining controlled conditions for initial validation. Future work will extend these evaluations to more complex layouts, dynamic obstacles, and a broader range of sensor noise levels.

A first experiment, Figure \ref{fig:all_states}, shows the three main stages in the evolution of the scene graph within the WM. Red circles denote interest zones referenced in the text.

\begin{itemize}
    \item Room acquisition: Following the sequences shown in Figure \ref{fig:all_states} as red ellipses, in Zone 1, the robot lacks a room representation. The $\mathcal{C}_{room}$ agent is aware of the absence of a room node and starts the initialisation process. This is reflected in the WM by the appearance of the $has\_intention$ link between the robot \textit{Shadow} and \textit{room\_measured}, Zone 2. The $\mathcal{C}_{room}$ waits until the \textit{mission\_monitoring} agent authorises the action.

    When the robot reaches the room's centre or sufficient data have been gathered, the $\mathcal{C}_{room}$ agent adds the room to the graph. The coordinate frame changes during this transition, and the robot’s pose is expressed relative to the room’s reference frame. This relationship is captured by the $RT$ link between the room and the robot, as shown in Zone 3. From now on, this link is updated by the \textit{energy\_optimiser} agent, which performs continuous optimisation over the room’s corner observations. 

     \item Door acquisition and affordance execution: With a current room established in WM, the $\mathcal{C}_{door}$ agent can now proceed with detecting doors. Zone 4 shows a new door proposal \textit{door\_3\_0\_1\_pre} and an action proposal in the edge \textit{has\_intention}. The robot moves close to the door, and the agent changes the door status from provisional to acquired, removing the \textit{pre} suffix. Additionally, the agent inserts an affordance node \textit{aff\_cross\_3\_0\_1} hanging from the new door to notify the \textit{mission\_monitoring} of that action's availability. This is shown in Zone 5.

     \item New room acquisition and door matching: Executing the door-crossing affordance, Zone 6, takes the robot into a new room, triggering a second initialisation process. Upon completion, the $\mathcal{C}_{room}$ agent inserts a new room node, \textit{room\_2}, and its constituent elements into Zone 7 of the graph. If the room had been previously known, the \textit{LTSM} agent would instead load the stored room, bypassing the initialisation step. Zone 8 illustrates how doors are associated across rooms immediately before the \textit{LTSM} agent removes the previously visited room and sets the new one as the current. \textcolor{black}{At this point, the door gets the dual coordinates that place it in both rooms. The \textit{LTSM} agent maintains a local graph of all known spaces and the open transitions between them, while adding and removing rooms as the robot explores its environment.}
\end{itemize}

A second experiment demonstrates a more extended activity within a ten-room scenario. \textcolor{black}{Following the steps described in the previous experiment, the robot incrementally builds its local and global representations and, at some point, closes a loop connecting several rooms.} A video of the full experiment can be downloaded from this site \footnote{https://cloudiepcc.unex.es/index.php/s/Zq2eY9fLeDa7itP}. All agents and the components in the sub-cognitive module run in different cores of the onboard computer, an Intel i9 13th generation processor with an NVidia RTX-A2000 GPU, and can sustain the nominal sensor acquisition rates, 20Hz for the 3D LiDAR and 30 Hz for the 360$^\circ$ camera.

\par

\begin{figure}[h!]
  \centering
  \subfloat[Room A and B\label{fig:room1}]{%
    \includegraphics[width=\textwidth]{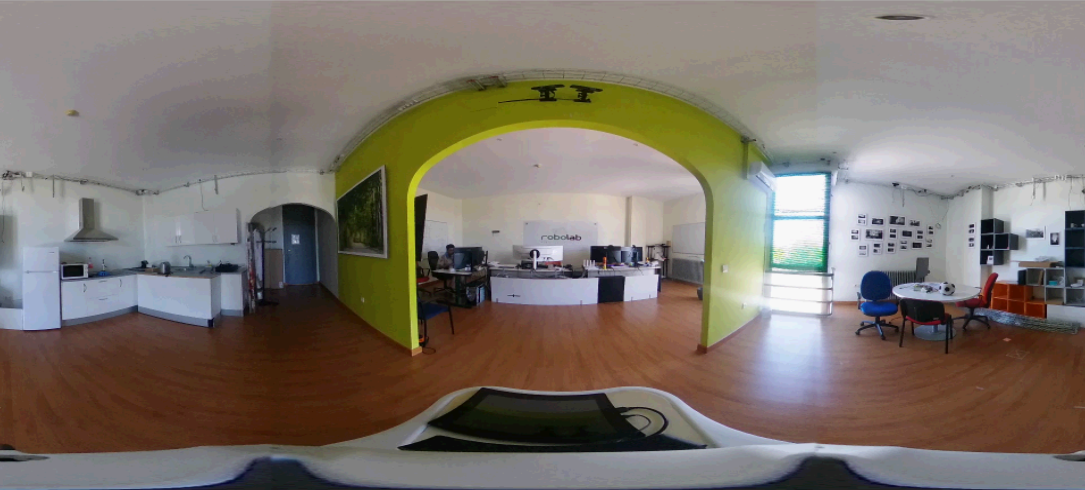}%
  }

  \vspace{0.3cm}

  \subfloat[Room A layout\label{fig:room1layout}]{%
    \includegraphics[width=0.45\textwidth]{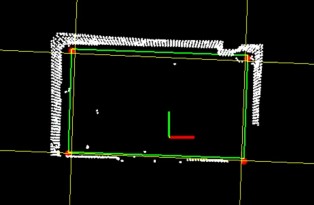}%
  }\hfill
  \subfloat[Room B layout\label{fig:room2layout}]{%
    \includegraphics[width=0.45\textwidth]{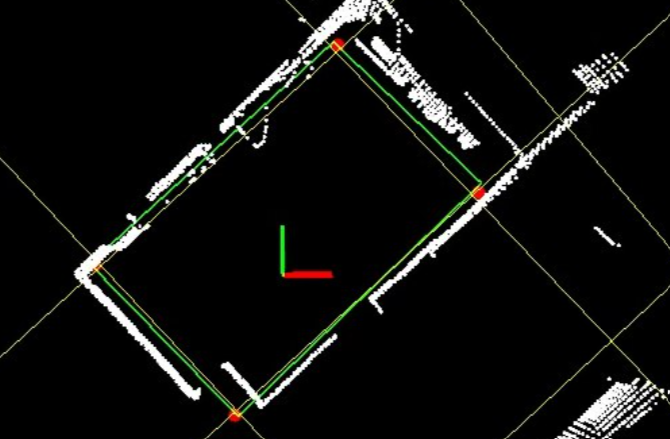}%
  }

  \caption{Real test scenario with their respective layouts.}
  \label{fig:rooms}
\end{figure}

The third experiment places the robot Shadow in a real scenario, navigating between two rooms separated by a door. Figure \ref{fig:rooms} shows the two rooms with their usual furniture and the room layouts extracted from them. Note that processing only the upper part of the 3D LiDAR field, $z>170cm$, eliminates most of the non-wall points, and the large structures detected with the Hough lines provide enough information to complete the correct layout. We are exploring ways of removing more of the remaining points that do not belong to the walls, for instance, by filtering the 3D point cloud with a semantic segmentation DNN. The robot has remained located with respect to the rooms' frames for four-hours navigation sessions, with maximum positioning errors of 15cm.

% --- Fila 1 ---

\subsection{\textcolor{black}{Measurement error analysis}}

\begin{figure}[h!]
    \centering
    \includegraphics[width=1\linewidth]{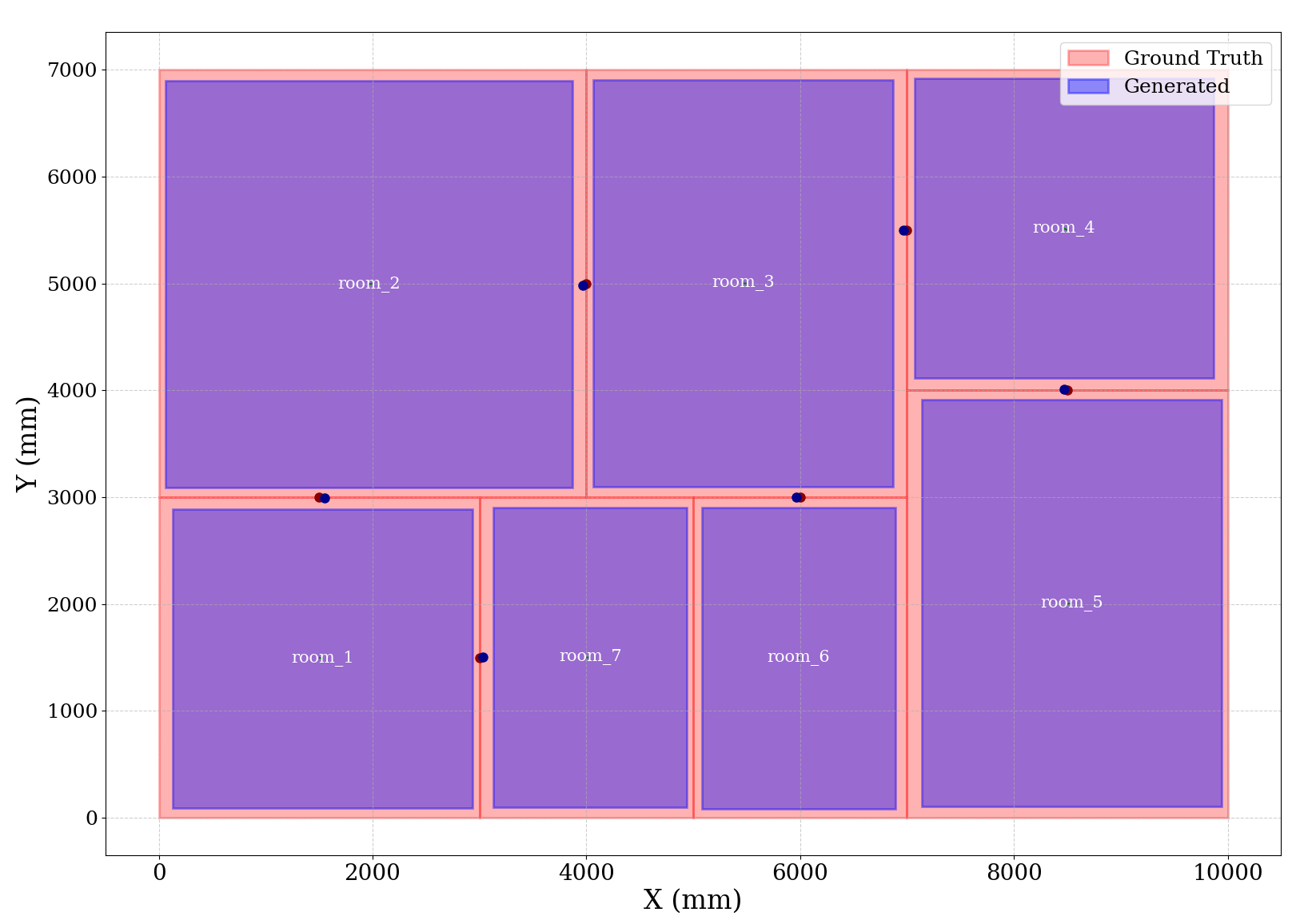}
    \caption{\textcolor{black}{Visual comparison between the ground truth floor plan and a generated result after the automatic alignment process. The ground truth is shown in red, while the generated layout is shown in violet. Rooms are represented as rectangles, and doors are marked with circles. The dashed green lines connect the centroids of the algorithmically matched rooms to indicate their correspondence.}}
    \label{fig:metric_results}
\end{figure}

\textcolor{black}{To quantitatively assess the geometric accuracy of the generated floor plans, a metric in both rooms and door widths was also negligible, demonstrating the system's adherence to the simulator. A visual representation of this superposition for one of the test runs can be seen in Figure \ref{fig:metric_results}.}

\textcolor{black}{The aggregated results from multiple generation runs are summarised in Table \ref{tab:error_summary}. The results indicate strong positional and dimensional fidelity. The low mean error in room placement suggests that the generated global map effectively maintains the correct spatial relationships among rooms. The standard deviation for this metric was minimal, suggesting high consistency and low variance across different outputs. Dimensional errors for both rooms and door widths were also negligible, demonstrating the system's ability to adhere to the specified architectural constraints. For the room dimensions, the errors are correlated with the wall thickness of the scenario, which is 200 mm. To prevent the propagation of wall-thickness–related errors, this factor is explicitly accounted for during the generation of the global map. These quantitative findings confirm the qualitative assessment, demonstrating that the generation model produces geometrically sound, accurate floor plans with minimal deviation from the target structure.}

\textcolor{black}{It is crucial to highlight that while the quantitative analysis reveals marginal geometric deviations, the most relevant outcome of this evaluation is the consistency and correctness in the construction of the topological map. Despite the minor positional and dimensional errors detailed in Table \ref{tab:error_summary}, the system successfully identified the correct topological structure, that is, which rooms connect to each other and through which doors, in all of the test cases. This success in inferring connectivity is fundamental to the robot's navigation and planning tasks, validating that the proposed approach generates environment representations that are not only geometrically accurate but also, more importantly, functionally and topologically reliable.}

\begin{table}[h!]
\centering
\caption{\textcolor{black}{Error Analysis Summary}}
\label{tab:error_summary}

\begin{tabular}{|l|c|c|}
\hline
\textbf{Metric} & \textbf{Mean error (mm)} & \textbf{Standard deviation (mm)} \\

\hline
Room position error & 34.1 & 16.2 \\
\hline
Room dimension error & 201.5 & 8.8 \\
\hline
Door position error & 37.5 & 13.9 \\
\hline
Door width error & 17.4 & 16.7 \\
\hline
\end{tabular}
\end{table}

\subsection{\textcolor{black}{Computational performance analysis}}

\begin{figure}[h!]
    \centering
    \includegraphics[width=1\linewidth]{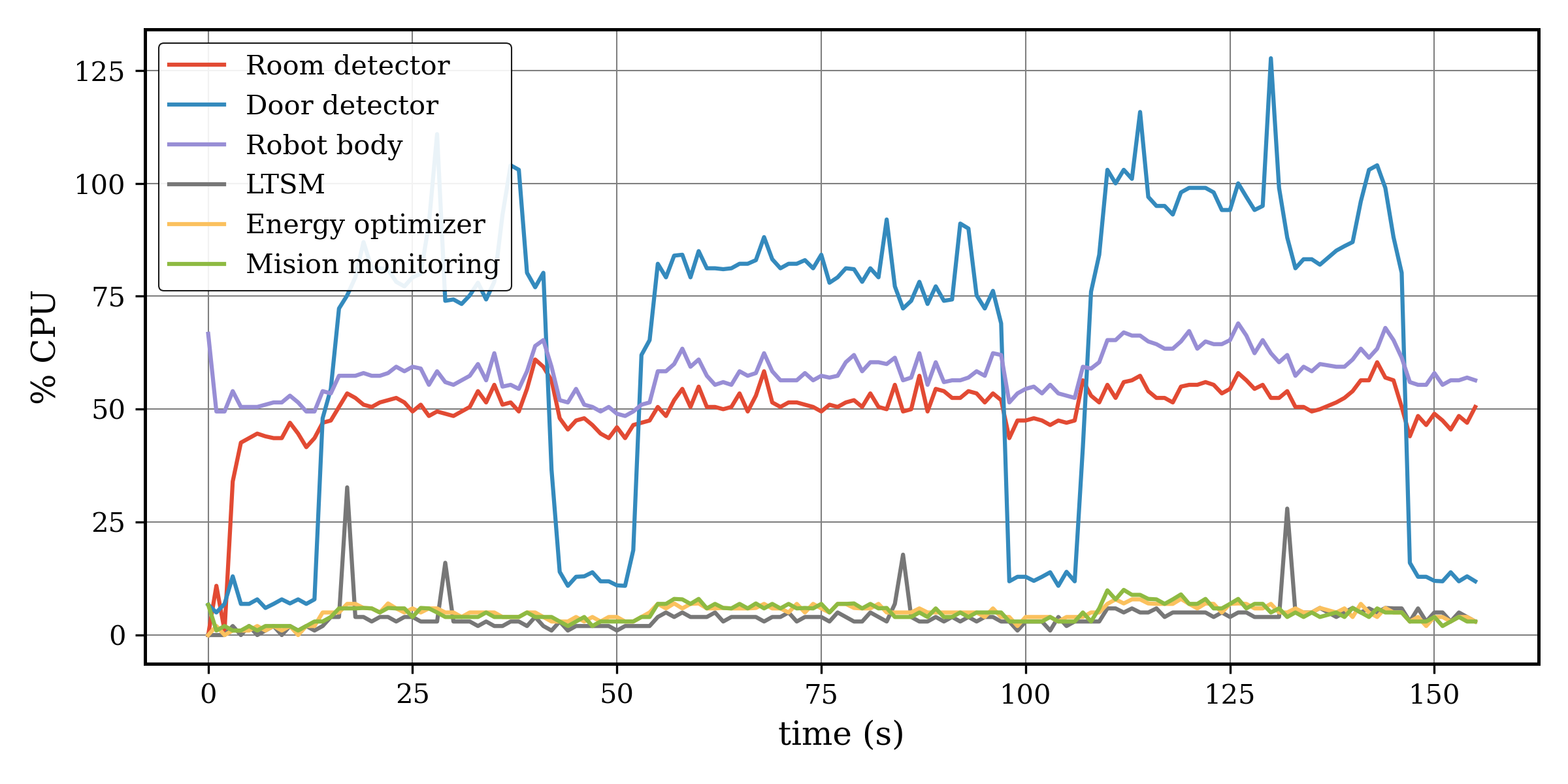}
    \caption{\textcolor{black}{CPU usage over time for the main processes running during the graph construction process.}}
    \label{fig:placeholder}
\end{figure}

\textcolor{black}{This section analyzes the computational performance of the architecture, focusing on the CPU and memory consumption of the main processes. The data was recorded during an experimental run that included the construction of a scene graph across several rooms, providing insight into the system's operational efficiency and resource allocation.
As shown in Figure \ref{fig:placeholder}, the CPU usage for most agents that constitute the architecture remains stable over time. The graph highlights the activation of the door detection process, marked by a notable increase in its CPU consumption, which occurs once a room model has been stabilised. This increased load is attributed to the agent's task of managing and validating multiple door instances within the active room.}

% \begingroup
% \color{blue}
\begin{table}[h!]
\centering
\caption{\textcolor{black}{Resource usage summary: Memory (Total system RAM: 64 GB)}}
\label{tab:ram_summary}

\begin{tabular}{|l|c|c|c|}
\hline
\textbf{Process} & \textbf{Mean RAM usage (\%)} & \textbf{Mean RAM usage (MB)} & \textbf{Std. Dev. (\%)} \\
\hline
Room detector & 0.30 & 197 & 0.00 \\
Door detector & 0.28 & 184 & 0.04 \\
Robot body & 0.31 & 203 & 0.03 \\
LTSM & 0.33 & 216 & 0.05 \\
Energy optimizer & 0.22 & 144 & 0.04 \\
Mision monitoring & 0.22 & 144 & 0.04 \\
\hline
\end{tabular}
\end{table}
%\endgroup

\textcolor{black}{Table \ref{tab:ram_summary} summarizes the memory consumption for the main architectural processes. The results demonstrate the system's high efficiency, with all agents operating with a minimal memory footprint. The mean RAM usage for each process is consistently low, ranging from 0.22\% to 0.33\%. Furthermore, the low standard deviation across all processes indicates stable and predictable memory consumption, a crucial feature for long-term autonomous operation on resource-constrained robotic platforms.}

\section{Noise Sensitivity and Detection Mistakes}

The robustness of concept-first spatial representation depends critically on reliable feature detection and consistent model validation. Two primary failure modes affect the current implementation: sensitivity to sensor noise and detection mistakes that manifest either immediately or after additional exploration.

\subsection{Noise Sensitivity}

\textcolor{black}{The quality of the final spatial representation degrades as sensor and odometry noise increase.} While systematic evaluation of noise sensitivity requires extensive parametric studies (planned for future work), preliminary observations from our simulations reveal several noise-related challenges. LiDAR noise affects corner detection reliability, particularly when points near wall intersections are displaced beyond the algorithm's clustering threshold. Similarly, odometry drift compounds during the initialisation phase, potentially causing misalignment between the selected corner set and the actual room geometry. \textcolor{black}{The GTSAM optimisation partially mitigates these effects by distributing error across the pose graph, but at high noise levels, incorrect room models can be accepted as valid.}

Future work will quantify these relationships through controlled experiments varying: (i) LiDAR point cloud noise ($\sigma = 0.01$m to $0.1$m), (ii) odometry drift rates ($0.5^\circ$ to $5^\circ$ of distance travelled), and (iii) angular measurement uncertainty ($0.5^\circ$ to $5^\circ$). The expected outcome is a noise tolerance envelope within which the concept-first approach maintains acceptable spatial accuracy.

\subsection{Detection Mistakes and Model Revision}

The current lifecycle implementation assumes that once validated, concept instances remain fixed. This rigid approach fails to handle two critical scenarios that occur in practice:

Early Model Mistakes: Despite validation, incorrect models may still satisfy the instantiation criteria due to partial occlusions, symmetric ambiguities, or sensor limitations. For example, a large L-shaped space might initially be interpreted as a rectangular room when only one branch is visible. The current architecture lacks mechanisms to revise these early commitments when contradictory evidence emerges.

Temporal Model Invalidation: Initially correct models may become inconsistent as exploration reveals previously unobserved features. A room model validated from limited viewpoints might prove incompatible with newly discovered walls or openings. Without dynamic revision capabilities, these inconsistencies accumulate, degrading the overall representation quality.

\textcolor{black}{Addressing these limitations requires extending the concept lifecycle with continuous model fitness evaluation and revision mechanisms. This would provide the corrective capabilities that the current system lacks. We envision a more flexible lifecycle incorporating:}

\begin{enumerate}
\item Continuous Fitness Monitoring: Each concept instance maintains a running fitness metric quantifying the agreement between predicted and observed features. \textcolor{black}{A persistent low fitness score would serve as the trigger to invalidate the model and its child concepts, forcing a re-evaluation.} This metric would track both recent observations (for rapid response) and historical consistency (for stability).

\item Graduated Response Strategy: When fitness degrades below threshold, the system would engage a hierarchical response: (i) local parameter adjustment for minor discrepancies, (ii) structural revision for significant geometric changes, and (iii) complete model replacement when the current instance becomes untenable.

\item Backtracking and Alternative Hypotheses: The architecture would maintain alternative concept instantiations as latent hypotheses, enabling rapid switching when the primary model fails. This requires extending the concept-agents to support probabilistic beliefs over multiple competing interpretations. A particle filter would be a good starting point.

\item Graceful Degradation: When no satisfactory model exists, the system should maintain partial representations rather than forcing incorrect instantiations. This might involve temporary metric patches or undefined regions marked for future exploration.
\end{enumerate}

These extensions would transform the current deterministic lifecycle into a probabilistic framework where concept instances compete, evolve, and occasionally fail. While computationally more demanding, this flexibility is essential for robust operation in real-world environments where perfect sensing and complete observability cannot be guaranteed.

\section{\textcolor{black}{Discussion}}

\textcolor{black}{The experimental results confirm the viability of our concept-first architecture for building actionable scene graphs. \textcolor{black}{Due to the fundamental architectural differences}, a direct quantitative comparison with existing metric-first systems is not our primary focus. Instead, this section contextualises our contributions and limitations by qualitatively comparing our approach against leading alternative frameworks.}

\textcolor{black}{Our architecture's primary advantage is its independence from a pre-existing global metric map, which contrasts with frameworks like Hydra that ground their 3D scene graph (3DSG) in a dense mesh. By building a world model from connected local reference frames, \textcolor{black}{our system can offer greater scalability while avoiding the need to maintain a single, monolithic metric representation.} However, a key trade-off of this design is that our sparse, concept-based representation lacks the dense geometric fidelity of metric-first systems. While this can be a disadvantage for low-level tasks such as navigation and collision avoidance in cluttered, unstructured environments, it's important to note that a metric grid can be created for each room when necessary to support these functions.}

\textcolor{black}{Our work prioritises \textcolor{black}{explicit modelling of environmental structure—such as rooms and walls—} as first-class entities in the graph. This enables our core principle of hierarchical constraint propagation—using the geometric context of a room to guide the search for a door—a top-down mechanism distinct from the primarily bottom-up, object-centric approach of many open-vocabulary systems. The clear drawback is our reliance on pre-defined concepts, which limits our semantic flexibility compared to open-vocabulary systems that can perceive and ground an open set of objects and relationships. However, this intentional design choice ensures that our architecture remains deterministic and interpretable. In contrast to the "black-box" nature of deep learning-based detectors used in many open-vocabulary systems, our approach provides clear, explicit mechanisms for reasoning about the environment's structure.}

\section{Conclusions and future works}

% This work presents an architectural exploration of concept-first spatial representation for mobile robots. By implementing room and door concepts as asynchronous agents within the CORTEX cognitive architecture, we demonstrate that semantic scene graphs can be built incrementally without requiring prior metric maps. The key innovation is hierarchical constraint propagation, where room instantiation provides geometric and semantic priors that improve door detection efficiency and robustness compared to unconstrained metric analysis.
% While our current implementation is limited to rectangular rooms in structured environments, the architectural principles—asynchronous concept agents, direct scene graph construction, and hierarchical constraint propagation—extend naturally to more complex spatial concepts and irregular geometries. The use of simple detection algorithms (Hough transforms, gap analysis) reflects our focus on architectural rather than algorithmic contributions; more sophisticated perception methods would enhance system performance while preserving the core organisational principles.

This work presents an architectural exploration of concept-first spatial representation for mobile robots. By implementing room and door concepts as asynchronous agents within the CORTEX cognitive architecture, we demonstrate that semantic scene graphs can be built incrementally without requiring prior metric maps. The key innovation is hierarchical constraint propagation, where room instantiation provides geometric and semantic priors that improve door detection efficiency and robustness compared to unconstrained metric analysis.
While our current implementation is limited to rectangular rooms in structured environments, the architectural principles —such as asynchronous concept agents, direct scene graph construction, and hierarchical constraint propagation—extend naturally to more complex spatial concepts and irregular geometries. \textcolor{black}{We intentionally employ simple detection algorithms (e.g., Hough transforms and gap analysis) to emphasise architectural rather than algorithmic contributions.} More sophisticated perception methods would enhance system performance while preserving the core organisational principles.

\textcolor{black}{An important avenue for future research is to extend the architecture to multi-robot or multi-agent scenarios.} While our current work is scoped to a single robot, a shared WM and LTSM across multiple agents could, in principle, enhance global graph consistency by exploiting inter-robot detections as additional constraints. This would help mitigate aggregated sensor noise, but also raises significant challenges in data association (e.g., ensuring that one robot’s $'room\_5'$ is correctly matched with another’s $'room\_2'$), communication latency, and distributed optimisation under heterogeneous sensing conditions. Addressing these challenges will be key to making the architecture scalable to cooperative multi-robot exploration.

This work leaves many open research topics, most of which are already under way: a) extending the system to handle non-rectangular rooms; b) replacing room and door detection algorithms with more robust data-driven learnt functions; c) using differentiable programming in the characterisation of object geometries so they can be adjusted online through the minimisation of a cost function; d) incorporating additional concept classes (e.g., furniture, household objects) into the scene graph; e) and implementing a more robust loop closure mechanism based on visual descriptors and objects stored within rooms.

% f) validating the system extensively on physical robot platforms in diverse real-world indoor environments, assessing performance beyond simulation; g) integrating the CORTEX semantic memory (an ontology) to provide on-the-fly common-sense enhancement of grounded objects; and h) extending the framework to handle dynamic elements and interactions with humans. 

The final goal is to demonstrate that concept-first architectures can provide transparent, human-interpretable spatial representations that support both autonomous robot operation and effective human-robot collaboration in real-world environments. This exploration represents an initial step toward cognitive architectures that reason about space using human-meaningful concepts rather than metric primitives, potentially enabling more natural and explainable robotic spatial intelligence.

\vspace{6pt} 

\section*{Author Contributions}
Conceptualization, P.B., N.Z., G.P.; methodology, N.Z. and G.P.; software, G.P., A.T. and N.Z.; validation, G.P., N.Z., and P.B.; formal analysis, N.Z. and P.N.; investigation, N.Z.; resources, N.Z.; data curation, N.Z. and G.P.; writing---original draft preparation, P.B., A.T., N.Z. and G.P.; writing---review and editing, P.N. and P.B.; visualization, A.T.; supervision, P.N. and P.B.; project administration, P.N.; funding acquisition, P.B. and P.N.; All authors have read and agreed to the published version of the manuscript.

\section*{Funding}
This work has been partially funded by FEDER Project 0124\_EUROAGE\_MAS\_4\_E (2021-2027 POCTEP Program) and by the Spanish Ministry of Science and Innovation PID2022-137344OB-C31 funded by MCIN/AEI/10.13039/501100011033/FEDER, UE. We want to express our gratitude to Lucas Bonilla for his outstanding work in the implementation of the system.

\section*{Institutional Review Board Statement}
Not applicable

\section*{Informed Consent Statement}
Not applicable

\section*{Acknowledgments}
The authors acknowledge the support described in the funding statement and the contributions of collaborators to the implementation of the system.

\section*{Conflicts of Interest}
The authors declare no conflicts of interest.

\end{document}